\documentclass[10pt]{article} % For LaTeX2e
\usepackage[accepted]{tmlr}
\usepackage{amsmath,amsfonts,bm}

\def\eqref#1{equation~\ref{#1}}
\def\1{\bm{1}}

\def\eps{{\epsilon}}

\DeclareMathAlphabet{\mathsfit}{\encodingdefault}{\sfdefault}{m}{sl}
\SetMathAlphabet{\mathsfit}{bold}{\encodingdefault}{\sfdefault}{bx}{n}

\newcommand{\Var}{\mathrm{Var}}

\usepackage{hyperref}
\usepackage{url}

\usepackage[utf8]{inputenc} % allow utf-8 input
\usepackage[T1]{fontenc}    % use 8-bit T1 fonts
\usepackage{hyperref}       % hyperlinks
\usepackage{url}            % simple URL typesetting
\expandafter\def\expandafter\UrlBreaks\expandafter{\UrlBreaks\do\-}           % break urls at -
\usepackage{booktabs}       % professional-quality tables
\usepackage{amsfonts}       % blackboard math symbols
\usepackage{nicefrac}       % compact symbols for 1/2, etc.
\usepackage{microtype}      % microtypography
\usepackage{xcolor}         % colors

\usepackage{enumerate}

\usepackage{amsthm}
\usepackage{amsmath, mathtools}

\usepackage{mleftright}

\usepackage{float}

\usepackage{threeparttable}
\usepackage{hhline}
\usepackage{multirow}

\usepackage{natbib}

\let\AND\relax
\usepackage{algorithm}
\usepackage{algorithmic}
\usepackage{tcolorbox}
\usepackage{caption}

\usepackage[capitalise]{cleveref}
\usepackage{tikz}
\usepackage{subcaption}
\usetikzlibrary{matrix}
\tikzset{mymatrix/.style={matrix of nodes, nodes={scale=0.8}}}

\usepackage{graphicx}
\usepackage{amsmath}
\usepackage{amssymb}
\usepackage{booktabs}
\usepackage{microtype}

\usepackage{thmtools}
\usepackage{thm-restate}
\newtheorem{theorem}{Theorem}
\newtheorem{definition}{Definition}[section]

\newtheorem{lemma}{Lemma}[section]
\newenvironment{proofof}[1]{{\bf Proof of #1:  }}{\hfill\rule{2mm}{2mm}}

\renewcommand{\Var}{\textrm{Var}}

\newcommand{\AutoAdjust}[3]{\mathchoice{ \left #1 #2  \right #3}{#1 #2 #3}{#1 #2 #3}{#1 #2 #3} }
\newcommand{\InParentheses}[1]{\AutoAdjust{(}{#1}{)}}
\newcommand{\InBrackets}[1]{\AutoAdjust{[}{#1}{]}}
\newcommand{\Ex}[2][]{\operatorname{\mathbf E}_{#1}\InBrackets{#2}}

\renewcommand{\eps}{{\varepsilon}}

\renewcommand{\vec}[1]{\mathbf{#1}}

\title{Not All Rollouts are Useful: \\ Down-Sampling Rollouts in LLM Reinforcement Learning}

\author{%
\name Yixuan Even Xu\thanks{Equal contribution.}\email yixuanx@cs.cmu.edu\\%
\addr Carnegie Mellon University \vspace{8pt}\\%
\AND%
\name Yash Savani\footnotemark[1]\email ysavani@cs.cmu.edu\\%
\addr Carnegie Mellon University \vspace{8pt}\\%
\AND%
\name Fei Fang\email feif@cs.cmu.edu\\%
\addr Carnegie Mellon University \vspace{8pt}\\%
\AND%
\name J. Zico Kolter\email zkolter@cs.cmu.edu\\%
\addr Carnegie Mellon University
}
\def\month{04}  % Insert correct month for camera-ready version
\def\year{2026} % Insert correct year for camera-ready version
\def\openreview{\url{https://openreview.net/forum?id=MfHOmgqVXM}} % Insert correct link to OpenReview for camera-ready version

\begin{document}

\maketitle

\begin{abstract}
    Reinforcement learning with verifiable rewards (RLVR) has emerged as the leading approach for enhancing reasoning capabilities in large language models. However, it faces a fundamental compute and memory asymmetry: rollout generation is embarrassingly parallel and memory-light, whereas policy updates are communication-heavy and memory-intensive. To address this, we introduce \textbf{PODS} (\textbf{P}olicy \textbf{O}ptimization with \textbf{D}own-\textbf{S}ampling), which decouples rollout generation from policy updates by training only on a strategically selected subset of rollouts, maintaining learning quality while dramatically reducing update costs. We propose a principled subset selection criterion---\emph{max-variance down-sampling}---that maximizes the variance of reward in the selected subset, and provide an efficient $O(n\log n)$ implementation of this rule. Empirically, Group Relative Policy Optimization (GRPO) coupled with PODS achieves the peak test accuracy of vanilla GRPO at least $\mathbf{1.7\times}$ \textbf{faster} across the different reasoning benchmarks and hardware configurations we tested.
\end{abstract}

\section{Introduction}
\label{sec:introduction}

Reinforcement learning with verifiable rewards (RLVR) has driven recent breakthroughs in solving math problems, code generation, and general reasoning for large language models (LLMs) \citep{jaech2024openai, ziegler2019fine, ouyang2022training, stiennon2020learning}. RLVR algorithms such as Proximal Policy Optimization (PPO) \citep{schulman2017proximal} and Group Relative Policy Optimization (GRPO) \citep{shao2024deepseekmath} share a two-phase structure: an \emph{inference phase}, which generates rollouts given a prompt, and a \emph{policy-update phase}, which updates the model parameters using the rewards calculated on those rollouts.

\begin{figure}[t]
    \centering
    \includegraphics[width=\textwidth]{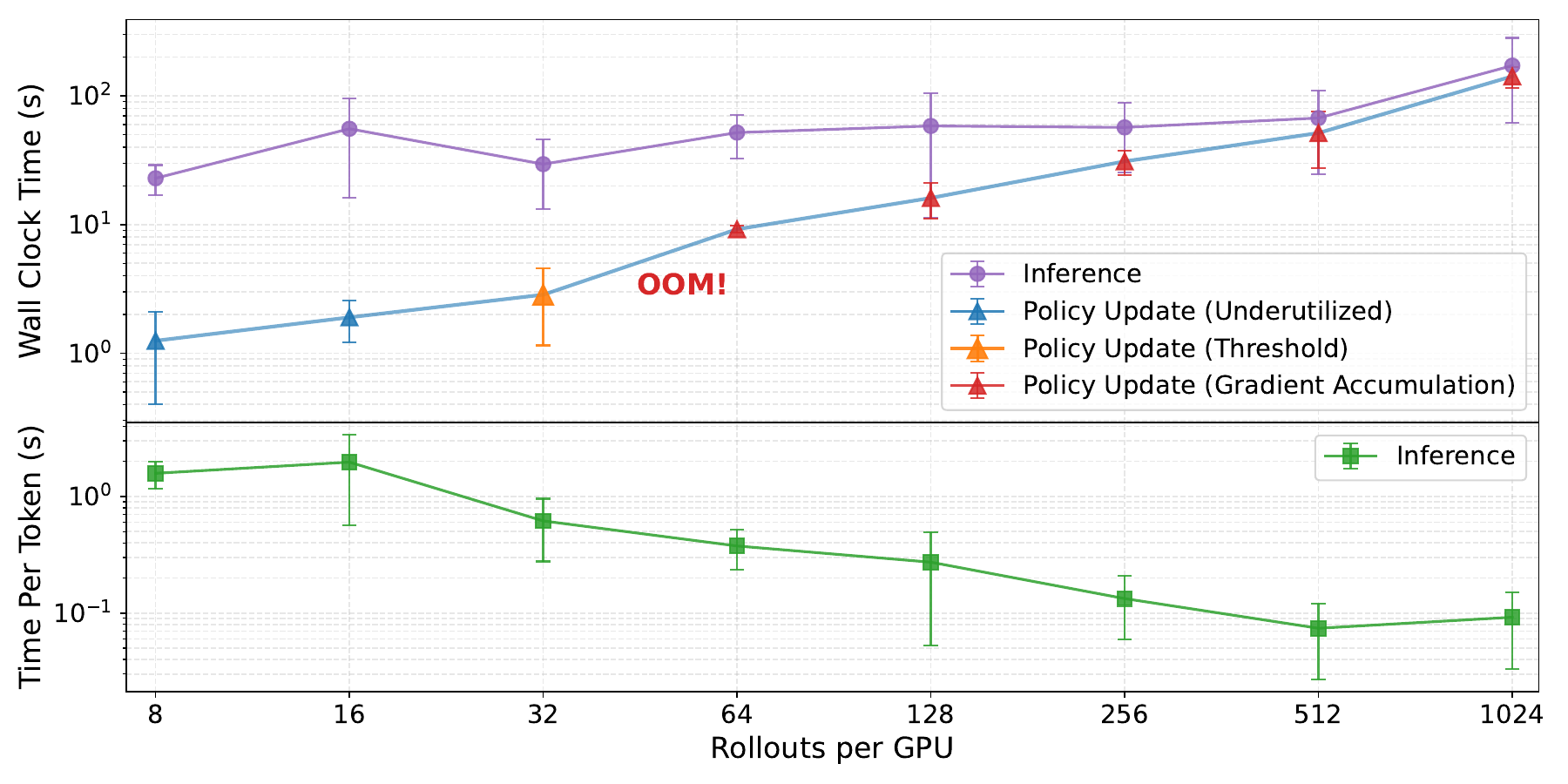}
    \caption{\textbf{Inference scales efficiently while policy updates become memory-bound in RLVR.} 
Empirical timing breakdown when fine-tuning Qwen2.5-3B-Instruct on GSM8K using $8$ A100-80GB GPUs with varying rollouts per GPU. 
\textbf{Top:} Total wall-clock time per iteration. Policy updates hit memory limits after $32$ rollouts per GPU (Out of memory beyond this point), requiring gradient accumulation that dramatically slows training. 
\textbf{Bottom:} Per-token inference time decreases $21\times$ through batching (from $8$ to $512$ rollouts), saturating beyond $512$. 
This demonstrates the core asymmetry that PODS exploits: inference parallelizes efficiently while policy updates become memory-bound.}
    % \caption{\textbf{Empirical breakdown of iteration time in RLVR.} We fine-tune Qwen2.5-3B-Instruct on GSM8K using $8$ A100-80GB GPUs, varying the number of rollouts per GPU. \textbf{Top:} Wall-clock iteration time for both inference and policy update. Policy update saturates memory at $32$ rollouts per GPU, beyond which larger batches require gradient accumulation and significantly slow iteration speed. While as we increase the number of rollouts per GPU, inference remains slower than policy update, but the \emph{marginal} increase in time with the scaling of batch size is dominated by the policy update. \textbf{Bottom:} Per-token inference latency is amortized by batching, yielding a $21\times$ speedup when increasing rollouts per GPU from $8$ to $512$, with scaling saturating near $512$. Together these panels highlight the inference-update asymmetry: inference parallelizes efficiently, while policy update becomes memory- and communication-bound beyond the $32$ rollouts per GPU threshold.}
    \label{fig:empirical}
\end{figure}

These two phases place different computational demands on the hardware. Inference is embarrassingly parallel and relatively memory-light, enabling modern accelerators to produce thousands of rollouts concurrently. Although generating a single rollout may have high latency due to autoregressive decoding, batching rollouts amortizes the per-token latency and yields higher throughput. Policy updates, on the other hand, scale poorly with batch size: they are memory- and communication-intensive, requiring full-precision optimizer states and cross-device synchronization of gradients and parameters. This asymmetry creates a fundamental bottleneck: systems must either throttle inference (underutilizing compute) or resort to memory-saving techniques like gradient accumulation (increasing communication overhead and policy update latency), both of which hurt training efficiency. \cref{fig:empirical} provides empirical evidence for this computational asymmetry.

\begin{figure}[t]
    \centering
    \includegraphics[width=\textwidth]{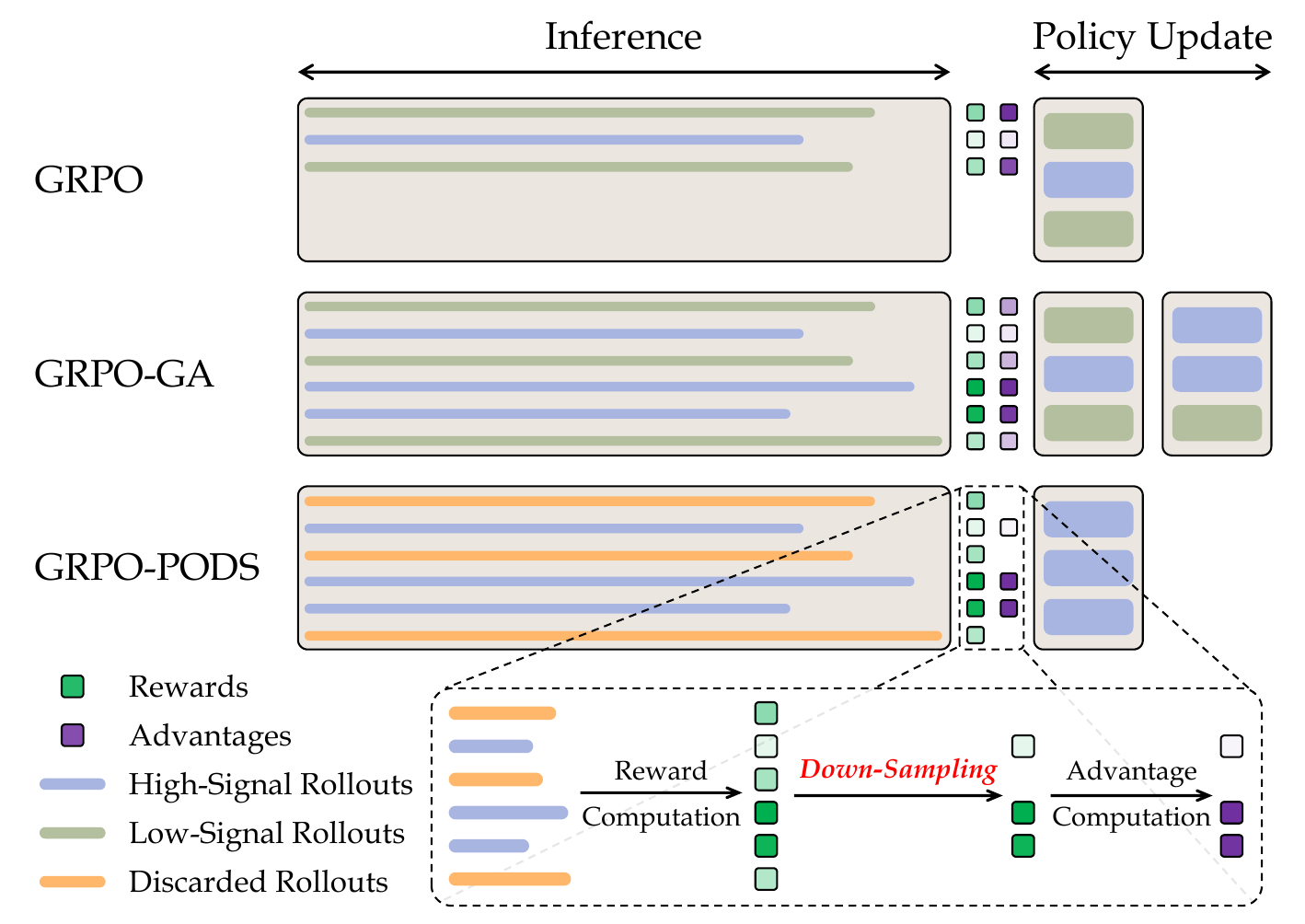}
    \caption{Visualization of three training strategies: vanilla GRPO, GRPO with gradient accumulation (GRPO-GA), and GRPO with PODS (GRPO-PODS). Vanilla GRPO generates $n$ rollouts and trains on all of them, leaving inference hardware underutilized due to the asymmetric computational demands of the two phases. GRPO-GA alleviates this issue with memory-saving techniques such as gradient accumulation, but at the cost of more sequential steps in the policy-update phase. In contrast, GRPO-PODS also generates $n$ rollouts but trains on only $m$ carefully selected examples, maximizing inference utilization, avoiding gradient-accumulation overhead, and providing a cleaner learning signal that yields better final performance.}
    \label{fig:method}
\end{figure}

A key observation of our work is: \emph{not all rollouts contribute equally to model improvement}. Beyond a certain scale, additional rollouts provide diminishing returns and can even degrade learning signals through redundant information. This suggests a natural solution: generate large batches of rollouts during the scalable inference phase, but train selectively on only the most informative subset during the policy update phase, avoiding the latency overhead of memory-saving techniques. We formalize this idea in \textbf{PODS} (\textbf{P}olicy \textbf{O}ptimization with \textbf{D}own-\textbf{S}ampling). As illustrated in \cref{fig:method}, PODS maximizes hardware utilization by generating $n$ rollouts per prompt but updating on only $m < n$ informative samples selected by a principled down-sampling rule.

% Our work provides a means to overcome this asymmetric bottleneck. We observe that \emph{not all rollouts contribute equally to model improvement}. Beyond a certain point, additional samples not only provide sharply diminishing returns, but also introduce redundant information into the training that can potentially blur the learning signal and hinder convergence. By training on a strategically chosen subset, we can recover the wall-clock time otherwise spent on superfluous policy updates while strengthening the learning signal. We formalize this idea in \textbf{PODS} (\textbf{P}olicy \textbf{O}ptimization with \textbf{D}own-\textbf{S}ampling). As illustrated in \cref{fig:method}, PODS maximizes hardware utilization during inference by generating a large pool of rollouts ($n$ per prompt) but updating the policy on only the $m < n$ informative examples selected by a principled down-sampling rule.

Within the PODS framework, we introduce \emph{max-variance down-sampling}, a principled criterion that selects the subset of rollouts with the greatest reward variance of the selected subset, thereby preserving strong contrastive signals. We show that the resulting combinatorial problem can be solved in $O(n \log n)$ time and, in the common binary-reward setting, reduces to picking the $m/2$ highest-reward and $m/2$ lowest-reward rollouts. We evaluate PODS with GRPO on GSM8K~\citep{cobbe2021training}, MATH~\citep{hendrycksmath2021} and the Chemistry subset of SciKnowEval~\citep{feng2024sciknoweval}  across multiple model and hardware configurations, demonstrating that it achieves the peak test accuracy of baseline GRPO at least $\mathbf{1.7\times}$ \textbf{faster}.

\section{Related Work}
\label{sec:related_work}

\paragraph{Reinforcement learning for LLM reasoning.}
Reinforcement learning has emerged as a powerful paradigm for enhancing the reasoning capabilities of LLMs across math, coding, and problem-solving domains \citep{jaech2024openai, shao2024deepseekmath, kazemnejad2024vineppo}. Although classical algorithms such as Proximal Policy Optimization (PPO) \citep{schulman2017proximal} laid the foundation, recent work has tailored them specifically to language models. Group Relative Policy Optimization (GRPO) \citep{shao2024deepseekmath} has gained prominence for reasoning tasks because of its implementation simplicity, competitive performance relative to PPO, and lack of a separate critic network. OpenAI o1 \citep{openai2025learning} and DeepSeek R1 \citep{guo2025deepseek}, which used large-scale RL, have sparked interest in reasoning-focused RL methods \citep{chen2025empirical, hu2025open, hu2025reinforce++, cui2025process}. Meanwhile, value-based approaches like PPO remain central \citep{yuan2025vapo, yuan2025s}, alongside complementary techniques such as Monte Carlo Tree Search \citep{gao2024interpretable, xie2024monte} and multi-agent methods \citep{meta2022human}. 
% Two recent works share superficial similarities to ours: DAPO \citep{yu2025dapo}, which dynamically skips problems with binary accuracy during training, and VAPO \citep{yuan2025vapo}, which advocates generating additional rollouts per problem instead of enlarging the batch size. However, neither method systematically \emph{down-samples} the rollouts produced during the inference-to-update pipeline—our key contribution. 
A recent line of work has also explored data selection for improving RL methods for LLM training. Specifically, prompt selection and filtering has gained significant attention from works such as DAPO \citep{yu2025dapo}, SRPO \citep{zhang2025srpo}, Reinforce-Rej \citep{xiong2025minimalist}, Polaris \citep{Polaris2025}, GRESO \citep{zheng2025act} and VAPO \citep{yuan2025vapo}. Our method advances this line of work by focusing on down-sampling rollouts within each prompt, instead of selecting or filtering prompts themselves. By tackling this computational-efficiency bottleneck, our approach complements existing methods and can be combined with them to further improve reasoning performance.

\paragraph{Down-sampling and data selection.}
The scale of modern machine learning necessitates effective data management strategies, particularly as datasets grow larger, noisier, and more imbalanced. Training on the full dataset can be prohibitively expensive, motivating sophisticated data-selection and down-sampling methods. Such techniques succeed across diverse settings—from theoretical results in clustering \citep{har2004coresets}, regression \citep{li2013iterative, rudelson2007sampling, clarkson2010coresets} to practical systems in speech recognition \citep{liu2015svitchboard, wei2014unsupervised} and computer vision \citep{kaushal2019learning, bankes2024reducr}. In reinforcement learning, prioritized experience replay \citep{schaul2015prioritized} and related methods \citep{hou2017novel, saglam2023actor, cusumano2025robust} highlight the value of selective sampling from experience buffers. More recently, careful data selection has become central to foundation-model training \citep{goyal2024scaling, schuhmann2021laion, gadre2023datacomp} and emerging applications such as computational advertising \citep{bei2023bidder, gravin2024bidder}. Yet, to our knowledge, we are the first to apply principled down-sampling to the rollout-generation stage of LLM reinforcement learning, mitigating a key computational bottleneck while strengthening the learning signal.

\section{Down-Sampling Rollouts in GRPO}
\label{sec:multi-grpo}

In this section, we present our approach to resolving the computational asymmetry between inference and policy updates in LLM reinforcement learning. We first review the original GRPO algorithm in \cref{subsec:preliminaries}, highlighting its \emph{structural} components and computational demands. Next, in \cref{subsec:multigrpo}, we introduce the \textbf{PODS} (\textbf{P}olicy \textbf{O}ptimization with \textbf{D}own-\textbf{S}ampling) framework, which strategically selects informative rollouts to maximize hardware utilization during both inference and policy-update phases. In \cref{subsec:maxvar}, we develop a principled \emph{max-variance down-sampling} method that preserves strong contrastive signals, justified by \citet{razin2025makes}, by retaining only rollouts from the extremes of the reward spectrum. We show that this method admits an elegant, $O(n\log n)$ solution, making it practical for real-world deployment. 
% Overall, our framework retains the advantages of GRPO while boosting computational and memory efficiency across diverse hardware setups.

\subsection{Preliminaries}
\label{subsec:preliminaries}

Group Relative Policy Optimization (GRPO) \citep{shao2024deepseekmath} is a reinforcement-learning algorithm intended to enhance the reasoning capabilities of large language models (LLMs), particularly within the RLVR setting. Each GRPO training step follows a structured, two-phase process, described below.

\paragraph{Inference phase.}
Let $\pi_{\theta}$ denote the policy parameterized by $\theta$, which defines a distribution over next-token probabilities given the previous tokens in a sequence. Given a single input prompt $p$ (e.g., a math problem), GRPO first generates a group of $n$ rollouts ${\vec{o}=(o_1, o_2, \dots, o_n)}$ by autoregressively sampling from $\pi_{\theta}$. Each rollout is a complete token sequence excluding the prompt, representing a possible solution. Each rollout is then evaluated using a reward model $r_i = R(o_i)$, which scores the quality and correctness of the corresponding output $o_i$. This yields a reward vector $\vec{r} = (r_1, r_2, \dots, r_n)$. We then compute normalized advantage estimates: $a_i = {(r_i - \mu)}/{\sigma}$, where $\mu$ and $\sigma$ are the mean and standard deviation of the rewards respectively.

\paragraph{Policy update phase.}
After computing the advantages, the policy is updated by optimizing the GRPO objective $L_{\mathrm{GRPO}}(\theta)$. Specifically, for each rollout $o_i$ with advantage $a_i$, we compute a loss for each token position $t$, and then average over all tokens and rollouts:
\begin{align*}
    L_{\mathrm{GRPO}}(\theta) = \ &\frac{1}{n} \sum_{i=1}^{n} \frac{1}{|o_i|} \sum_{t=1}^{|o_i|} \\
    &\min\left[
        \frac{\pi_{\theta}(o_{i,t} \mid p, o_{i,<t})}{\pi_{\theta_{\mathrm{fixed}}}(o_{i,t} \mid p, o_{i,<t})} \cdot a_i,\ 
        \mathrm{clip}\left(
            \frac{\pi_{\theta}(o_{i,t} \mid p, o_{i,<t})}{\pi_{\theta_{\mathrm{fixed}}}(o_{i,t} \mid p, o_{i,<t})}, 
            1 - \epsilon, 1 + \epsilon
        \right) \cdot a_i
    \right].
\end{align*}
where $|o_i|$ is the number of tokens in $o_i$ and $\pi_{\theta_{\mathrm{fixed}}}$ is a frozen copy of the policy used for importance weighting. This asymmetric loss embodies the \emph{slow to adopt, quick to abandon} learning principle—limiting how aggressively the policy increases probabilities for tokens in high-reward rollouts while allowing more substantial reductions for low-reward sequences.

\subsection{PODS Framework}
\label{subsec:multigrpo}

We propose to \emph{decouple the inference and training phases} in GRPO. Rather than updating on every generated rollout, PODS first produces \(n\) rollouts in parallel and then trains on only a smaller subset of size \(m < n\) selected by a down-sampling rule \(D\). This strategy exploits parallelism during inference while substantially reducing the communication and memory costs of the subsequent policy update.

\begin{definition}[Down-sampling rule]
    $D(\vec o, \vec r; m)$ is a function that takes $n$ rollouts $\vec o =~(o_1, o_2, \ldots, o_n)$, their corresponding rewards $\vec r = (r_1, r_2, \ldots, r_n)$, and the update size $m$. It outputs a subset of indices $S \subseteq \{1, 2, \dots, n\}$, where $|S|=m$, indicating which rollouts to retain for the policy update phase.
\end{definition}

Given a selected subset of indices $S$, we compute the advantage estimates using only the selected rollouts:  $a_{S, i} = (r_i - \mu_S) / \sigma_S$, where $\mu_S$ and $\sigma_S$ are the mean and standard deviation of the rewards in the selected subset. The GRPO-PODS objective then becomes:
\begin{align*}
    L_{\mathrm{PODS}}(\theta, S) = \ &\frac{1}{m} \sum_{i \in S} \frac{1}{|o_i|} \sum_{t=1}^{|o_i|} \\
    &\min\InBrackets{\frac{\pi_{\theta}(o_{i,t} \mid p, o_{i, <t})}{\pi_{\theta_{\mathrm{fixed}}}(o_{i,t} \mid p, o_{i, <t})} \cdot a_{S,i}, \text{clip}\InParentheses{\frac{\pi_{\theta}(o_{i,t} \mid p, o_{i, <t})}{\pi_{\theta_{\mathrm{fixed}}}(o_{i,t} \mid p, o_{i, <t})}, 1 - \eps, 1 + \eps} \cdot a_{S,i}}.
\end{align*}

\begin{algorithm}
    \caption{The PODS Framework for GRPO}
    \label{alg:multi-grpo}
    \begin{algorithmic}[1]
        \REQUIRE Models $\pi_\theta, \pi_{\theta_{\mathrm{fixed}}}$, input prompt $p$, reward model $R$,\\
        \hspace{1.35em} Number of rollouts $n$, update size $m$, down-sampling rule $D$
        \STATE Independently sample $n$ rollouts $\vec o = (o_1, o_2, \ldots, o_n)$ using $\pi_{\theta_{\mathrm{fixed}}}$ for prompt $p$
        \STATE Compute rewards $\vec r = (r_1, r_2, \ldots, r_n)$ using the reward model $R$
        \STATE Down-sample a set of $m$ rollouts $S \leftarrow D(\vec o, \vec r; m)$
        \STATE Update the policy using the GRPO-PODS objective $L_{\mathrm{PODS}}(\theta, S)$
        \ENSURE An updated model $\pi_{\theta_{\mathrm{updated}}}$
    \end{algorithmic}
\end{algorithm}

\Cref{alg:multi-grpo} outlines the PODS framework for GRPO with a single prompt $p$ in a training iteration. When training on a batch of multiple prompts, we simply apply the same procedure to each prompt and then concatenate the down-sampled rollouts and rewards. We conclude this section by presenting two trivial down-sampling strategies that can potentially be applied within PODS.

\paragraph{Random down-sampling.}
The rule $D_{\mathrm{rand}}$ uniformly selects $m$ indices from $\{1,2,\dots,n\}$ without replacement, thereby preserving the statistical properties of the original rollout distribution. In expectation, it yields the same parameter update as running standard GRPO on exactly $m$ rollouts.

\paragraph{Percentile down-sampling.} The rule $D_{\mathrm{perc}}$ selects the $m$ rollouts by choosing the $\frac{0.5}{m}$, $\frac{1.5}{m}$, $\dots$, $\frac{m-0.5}{m}$ quantiles of the reward distribution. This method ensures that the selected rollouts are evenly spaced across the reward spectrum, providing a more representative sample than random down-sampling.

\paragraph{Max-reward down-sampling.}
The rule \(D_{\mathrm{maxr}}\) selects the \(m\) rollouts with the highest rewards, concentrating on examples that exhibit the most desirable behavior. This should allow the model to learn primarily from successful reasoning patterns. However, as we show in \cref{sec:experiments}, ignoring low-reward rollouts deprives the policy of negative feedback and can significantly degrade performance.

\subsection{Max-Variance Down-Sampling}
\label{subsec:maxvar}

We now introduce \emph{max-variance down-sampling}, a principled down-sampling rule that selects the most diverse and informative rollouts according to their reward distribution.

Specifically, $D_{\mathrm{maxv}}$ chooses the subset $S$ of size $m$ that maximizes the empirical reward variance, i.e., $S=\arg\max_{|S|=m}\Var(\{r_i \mid i\in S\})$. By spanning the full performance spectrum, it supplies strong contrastive signals between successful and unsuccessful reasoning paths. Recent work by \citet{razin2025makes} provides an optimization-theoretic and empirical justification for this criterion. 

A naive search would examine $O\InParentheses{\binom{n}{m}}$ subsets. This is clearly infeasible for realistic $n$ and $m$. We prove, however, that the optimal subset can be found in $O(n\log n)$ time.

\begin{lemma}
    \label{lem:maxvar}
    For a sorted list of rewards $r_1 \leq r_2 \leq \cdots \leq r_n$, the variance-maximizing subset of size $m$ always consists of the $k$ highest rewards and $(m-k)$ lowest rewards for some $k \in \{0,1,\ldots,m\}$. That is,
    \begin{align*}
        \mathrm{Var}(\{r_1, \dots, r_{m-k}\} \cup \{r_{n-k+1},\dots, r_n\}) = \max_{|S| = m}\mathrm{Var}(\{r_i \mid i \in S\}).
    \end{align*}
\end{lemma}

\begin{proofof}{\cref{lem:maxvar}}
    Let $S^* = \arg\max_{|S| = m}\mathrm{Var}(\{r_i \mid i \in S\})$ be the optimal subset of size $m$. We will show that if $S^*$ is not of the form $\{1, \dots, {m-k}\} \cup \{{n-k+1},\dots, n\}$ for any $k$, then we can modify $S^*$ to obtain a new subset $S'$ of the same size with no smaller variance in rewards. By repeating this procedure, we can eventually reach a subset of this form.

    Let $\mu$ be the mean of the rewards in $S^*$. Since $S^*$ does not take the form of ${\{1, \dots, {m-k}\} \cup \{{n-k+1},\dots, n\}}$ for any $k$, there exists either (i) an element $i \in S^*$ such that $i>1, r_i \leq \mu$ and $i - 1 \not \in S^*$, or (ii) an element $j \in S^*$ such that $j<n, r_j \geq \mu$ and $j + 1 \not \in S^*$. That is, there exists an element in $S^*$, such that another element further from $\mu$ is not in $S^*$. We will show that we can swap them without decreasing variance.

    For the ease of notation, we will denote $\mathrm{Var}(\{r_i \mid i \in S\})$ as $\mathrm{Var}(S)$ in this proof.

    For case (i), let $S' = (S^* \setminus \{i\}) \cup \{i-1\}$, and let $\mu'$ be the mean of the rewards in $S'$. Then
    \begin{align*}
        \mathrm{Var}(S') - \mathrm{Var}(S^*) &= \InParentheses{\frac{1}{m}\sum_{t \in S'} r_t^2 - \mu'^2} - \InParentheses{\frac{1}{m}\sum_{t \in S^*} r_t^2 - \mu^2} \\
        &= \frac{1}{m}(r_{i-1}^2 - r_i^2) - (\mu'^2 - \mu^2) \\
        &= \frac{1}{m}(r_{i-1} - r_i)(r_{i-1} + r_i) - (\mu' - \mu)(\mu' + \mu) \\
        &= \frac{1}{m}(r_{i-1} - r_i)[(r_{i-1} + r_i) - (\mu' + \mu)] \geq 0.
    \end{align*}

    For case (ii), let $S' = (S^* \setminus \{j\}) \cup \{j+1\}$, we can similarly show that $\mathrm{Var}(S') - \mathrm{Var}(S^*) \geq 0$.

    In either case, we have shown that we can modify $S^*$ to obtain a new subset $S'$ of the same size that has no smaller variance in rewards. We can repeat this process until we reach a subset of the form $\{1, \dots, {m-k}\} \cup \{{n-k+1},\dots, n\}$ for some $k$. Thus, we conclude that there must exist one optimal subset of this form for some $k$.
\end{proofof}

\cref{lem:maxvar} naturally leads to a practical algorithm, \cref{alg:maxvar}, for max-variance down-sampling. Moreover, it also offers intuition as to why maximizing variance is effective: the optimal subset contains the $k$ highest and the $(m-k)$ lowest rewards, capturing contrastive signals from both positive and negative examples.

\begin{algorithm}
    \caption{Max-Variance Down-Sampling}
    \label{alg:maxvar}
    \begin{algorithmic}[1]
        \REQUIRE Number of rollouts $n$, update size $m$, rollouts $\{o_1, o_2, \ldots, o_n\}$, rewards $\{r_1, r_2, \ldots, r_n\}$
        \STATE Sort the rollouts by reward and get the sorted indices $ind \leftarrow \operatorname{argsort}(\{r_1, r_2, \ldots, r_n\})$
        \STATE Let $S_{\mathrm{ans}} \leftarrow \{ind_1, \dots, ind_{m}\}$
        \FOR{$k \in \{1, \dots, m\}$}
            \STATE Let $S_{\mathrm{this}} \leftarrow  \{ind_1, \dots, ind_{m-k}\} \cup \{ind_{n-k+1},\dots, ind_n\}$
            \STATE Let $S_{\mathrm{ans}} \leftarrow S_{\mathrm{this}}$ \textbf{if} $\mathrm{Var}(\{r_i \mid i \in S_{\mathrm{this}}\}) > \mathrm{Var}(\{r_i \mid i \in S_{\mathrm{ans}}\})$
        \ENDFOR
        \ENSURE Selected indices $S_{\mathrm{ans}}$ of rollouts
    \end{algorithmic}
\end{algorithm}

\begin{theorem}
    \label{thm:maxvar}
    \cref{alg:maxvar} computes the max-variance down-sampling rule correctly. Moreover, it can be implemented in $O(n \log n)$ time.
\end{theorem}

\begin{proofof}{\cref{thm:maxvar}}
    The correctness of \cref{alg:maxvar} follows directly from \cref{lem:maxvar}. 

    For the time complexity, we first sort the rewards in $O(n \log n)$ time. To compute the variance of the selected rollouts, note that $\Var(\{x \mid x \in S_{\text{this}}\}) = \Ex[x \in S_{\text{this}}]{x^2} - (\Ex[x \in S_{\text{this}}]{x})^2$. We can maintain the prefix sums of the rewards and the squared rewards in $O(n)$ time. Then, for each $k$, we can compute the variance of the selected rollouts in $O(1)$ time using the prefix sums. Thus, the overall time complexity is $O(n \log n) + O(m) = O(n \log n)$.
\end{proofof}

\cref{thm:maxvar} shows that the max-variance down-sampling rule can be computed efficiently, which enables its practical application in GRPO-PODS. We conclude this section by noting an important special case of the max-variance down-sampling rule.

\begin{theorem}
    \label{thm:maxvar_special}
    Let $m$ be an even integer. When the rewards are binary, selecting $m/2$ rollouts with the highest rewards and $m/2$ rollouts with the lowest rewards maximizes the variance of the rewards. 
\end{theorem}

\begin{proofof}{\cref{thm:maxvar_special}}
    Let the number of rollouts with reward 1 be $k$. Then, the number of rollouts with reward 0 is $n-k$. If $k \leq m/2$, then any subset of $m$ rollouts contains at most $k$ rollouts with reward 1, and the variance is maximized by selecting these $k$ rollouts and any $(m-k)$ rollouts with reward 0. If $n - k \leq m/2$, then any subset of $m$ rollouts contains at most $(n-k)$ rollouts with reward 0, and the variance is maximized by selecting these $(n-k)$ rollouts and any $m - (n - k)$ rollouts with reward 1. Otherwise, any subset of $m/2$ rollouts with reward 1 and $m/2$ rollouts with reward 0 maximizes the variance. In all cases, we can select $m/2$ rollouts with the highest rewards and $m/2$ rollouts with the lowest rewards to maximize the variance. This concludes the proof.
\end{proofof}

\section{Experiments}
\label{sec:experiments}

We evaluate PODS across diverse hardware configurations, model architectures, and model scales to demonstrate its generalizability and practical benefits. We test on three reasoning benchmarks of different domains, GSM8K \citep{cobbe2021training}, MATH \citep{hendrycksmath2021} and the L3 Chemistry subset of SciKnowEval \citep{feng2024sciknoweval}, using Qwen2.5 \citep{qwen2025qwen25technicalreport} and Llama3.2 \citep{meta2024llama3.2} models ranging from 3B to 7B parameters. Our experimental design covers both resource-constrained single-GPU setups and multi-GPU distributed training to validate PODS across different deployment scenarios. \cref{tab:experiments} describes our experimental configurations. We publish the code for our experiments at \href{https://github.com/YixuanEvenXu/pods}{https://github.com/YixuanEvenXu/pods}.

% We evaluate PODS with Qwen2.5 \citep{qwen2025qwen25technicalreport} and Llama3.2 \citep{meta2024llama3.2} on two reasoning benchmarks, GSM8K \citep{cobbe2021training} and MATH \citep{hendrycksmath2021}. Our experiments cover six hardware and model settings, as summarized in \cref{tab:experiments}.

\begin{table}[ht]
    \centering
    \begin{tabular}{cllccc}
        \toprule
        \textbf{Setting} & \textbf{Benchmark} & \textbf{Model} & \textbf{Parameters} & \textbf{GPUs} & \textbf{Fine-tuning Method} \\
        \midrule
        (a) & GSM8K     & Qwen2.5 & 3B   & 1 L40S & LoRA (rank 64, $\alpha=64$) \\
        (b) & GSM8K     & Llama3.2 & 3B  & 1 L40S & LoRA (rank 64, $\alpha=64$) \\
        (c) & MATH      & Qwen2.5 & 3B   & 1 L40S & LoRA (rank 64, $\alpha=64$) \\
        (d) & Chemistry & Qwen2.5 & 3B   & 1 L40S & LoRA (rank 64, $\alpha=64$) \\
        (e) & GSM8K     & Qwen2.5 & 3B   & 8 H100s & Full-Parameter \\
        (f) & GSM8K     & Qwen2.5 & 7B   & 8 A100s & Full-Parameter \\
        \bottomrule
    \end{tabular}
    \caption{Experimental configurations testing GRPO with PODS across task domains, model scales, hardware constraints, and training paradigms. Settings (a-d) test resource-constrained scenarios with LoRA fine-tuning, while (e-f) evaluate full-parameter training with distributed setups.}
    \label{tab:experiments}
\end{table}

\paragraph{Training infrastructure.} For settings (a-d), we use \texttt{Unsloth} \citep{unsloth} with \texttt{TRL} \citep{vonwerra2022trl} for efficient LoRA \citep{hu2022lora} reinforcement learning. For settings (e-f), we implement distributed training with \texttt{DeepSpeed ZeRO-2} \citep{rajbhandari2020zero} and extend the \texttt{open-r1} library \citep{openr1} to support PODS on multiple devices.

\paragraph{Rewards and evaluations.} We employ rule-based reward models that score rollouts, following standard practices in reasoning evaluation. Specifically, we reward an answer for correctness, format compliance, and the right number of thinking tags separately, resulting in a discrete but non-binary reward function. We provide additional details about the reward models \cref{subsec:rewards}.

% A rule-based reward model scores each rollout for correctness and format. 
% % Following \citep{yu2025dapo}, we omit the KL-divergence term from the GRPO objective because it is unnecessary for reasoning tasks. 
% For settings (d) and (e), the distributed training is implemented with \texttt{DeepSpeed ZeRO-2} \citep{rajbhandari2020zero}. We extend the \texttt{open-r1} library \citep{openr1} to support PODS on multiple GPUs. The detailed hyperparameters are listed in \cref{subsec:exp_hyperparams}. We adopt the same reward functions as \texttt{open-r1} \citep{openr1}. 
% % We publish the code for our experiments at \url{https://github.com/YixuanEvenXu/pods}.
% We have submitted our code to the supplementary materials and will release it upon publication of our work.

\paragraph{Section roadmap.} 

In \cref{subsec:exp_grpo}, we compare the performance of GRPO and GRPO-PODS across six hardware and model settings listed in \cref{tab:experiments}. We show that for all the settings we test, GRPO-PODS consistently outperforms GRPO in terms of test performance throughout the training. Then, in \cref{subsec:exp_params}, we focus on setting (a), and analyze the effect of the rollout and update sizes $(n,m)$ on the performance of GRPO-PODS, providing empirical insights into how to choose the rollout and update sizes for GRPO-PODS. We also present additional experiments about different down-sampling rules (\cref{subsec:exp_rules}) and different design choices of advantage normalization (\cref{subsec:exp_adv_norm}), additional evaluation results about PODS' speed up ratio compared to GRPO (\cref{subsec:exp_speedup}) and test performance generalization across different test sets (\cref{subsec:exp_additional_eval}), and the average response length over the course of training (\cref{subsec:exp_length}).

\begin{figure}[t!]
    \centering
    \begin{subfigure}{0.84\textwidth}
        \centering
        \includegraphics[width=\textwidth]{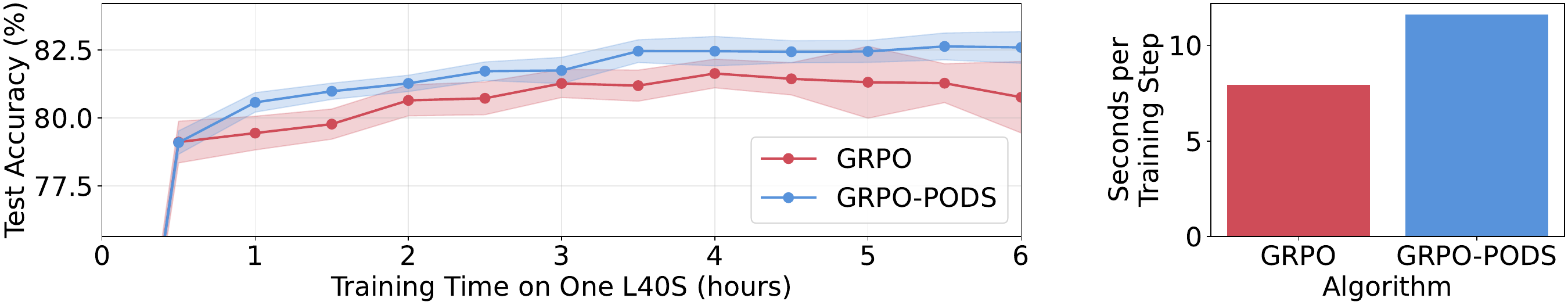}
        \caption{Training Qwen2.5 (3B) on GSM8K with one L40S GPU}
        \label{subfig:gsm8k_main}
    \end{subfigure}
    \hfill
    \begin{subfigure}{0.84\textwidth}
        \centering
        \includegraphics[width=\textwidth]{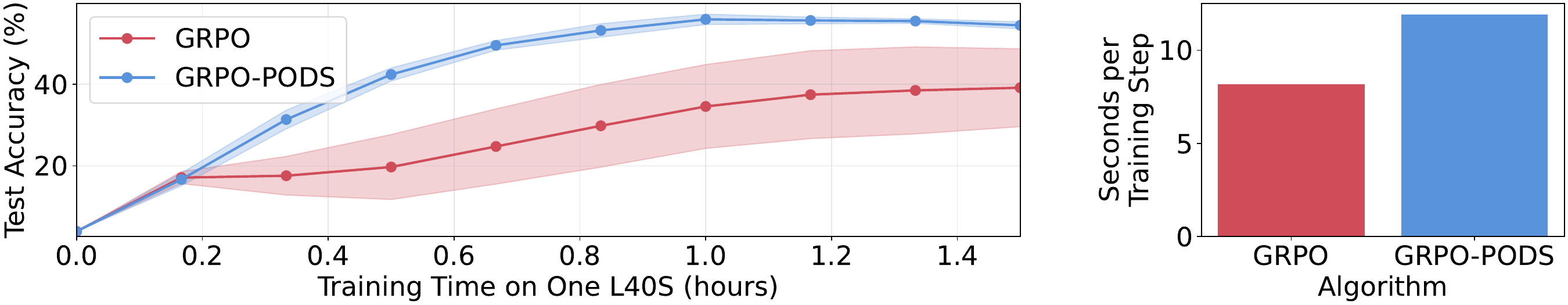}
        \caption{Training Llama3.2 (3B) on GSM8K with one L40S GPU}
        \label{subfig:llama_main}
    \end{subfigure}
    \hfill
    \begin{subfigure}{0.84\textwidth}
        \centering
        \includegraphics[width=\textwidth]{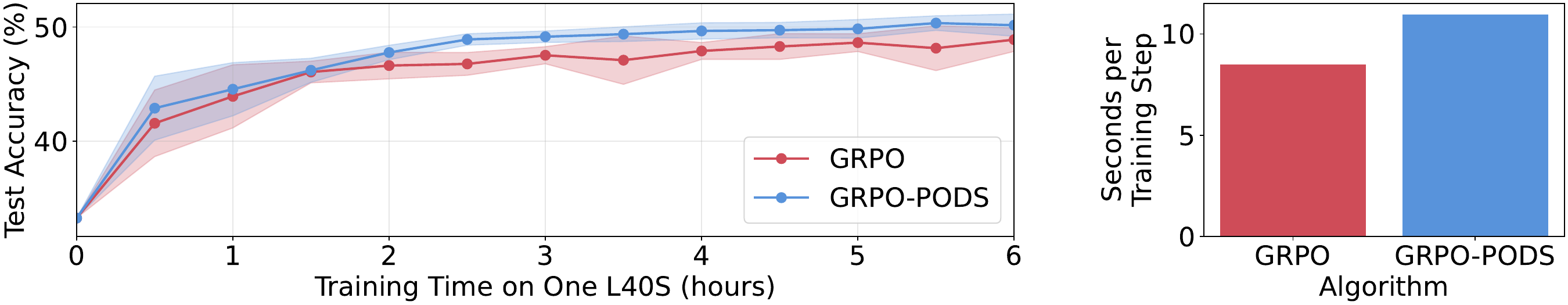}
        \caption{Training Qwen2.5 (3B) on MATH with one L40S GPU}
        \label{subfig:math_main}
    \end{subfigure}
    \hfill
    \begin{subfigure}{0.84\textwidth}
        \centering
        \includegraphics[width=\textwidth]{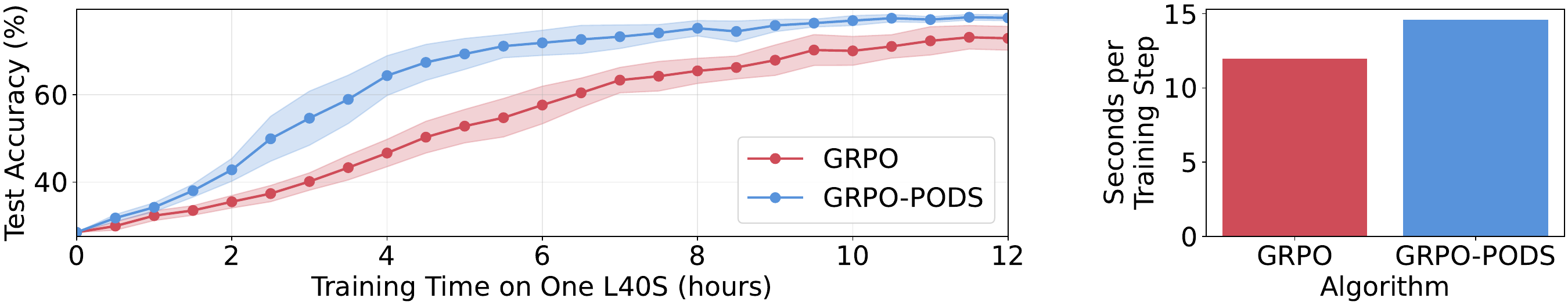}
        \caption{Training Qwen2.5 (3B) on Chemistry with one L40S GPU}
        \label{subfig:sciknow_main}
    \end{subfigure}
    \hfill
    \begin{subfigure}{0.84\textwidth}
        \centering
        \includegraphics[width=\textwidth]{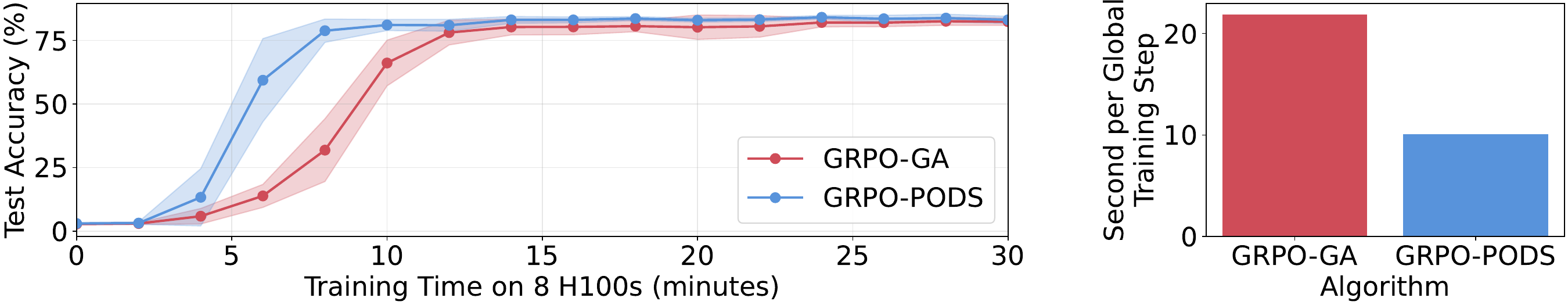}
        \caption{Training Qwen2.5 (3B) on GSM8K with 8 H100 GPUs}
        \label{subfig:gsm8k_scale}
    \end{subfigure}
    \hfill
    \begin{subfigure}{0.84\textwidth}
        \centering
        \includegraphics[width=\textwidth]{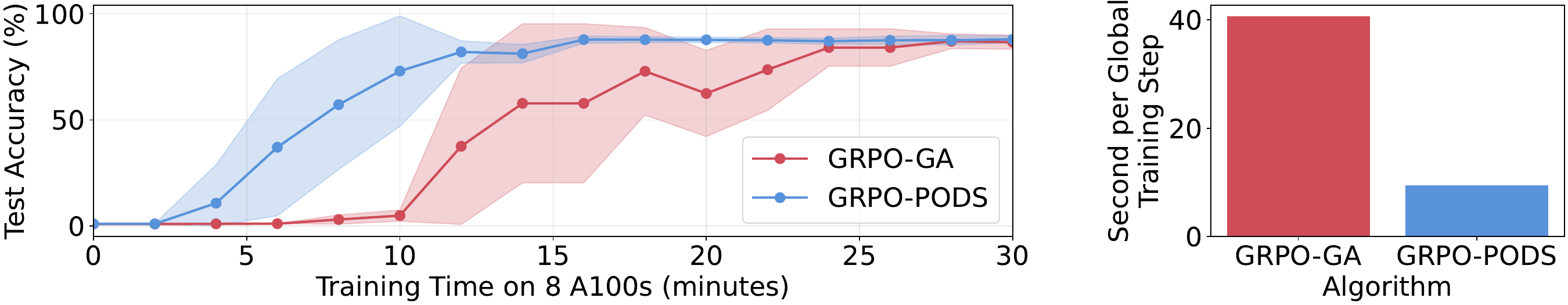}
        \caption{Training Qwen2.5 (7B) on GSM8K with 8 A100 GPUs}
        \label{subfig:gsm8k_scale_7b}
    \end{subfigure}
    \caption{Performance and per-step run time comparison of standard GRPO and GRPO-PODS with max-variance down-sampling across different datasets and hardware environments. For the performance comparison, the x-axis shows the training time, and the y-axis shows the accuracy on the test set. The shaded area represents 1.96 times the standard error of the mean.}
    \label{fig:main} 
\end{figure}

\subsection{Comparing GRPO-PODS to baseline GRPO}
\label{subsec:exp_grpo}

We evaluate max-variance down-sampling PODS with GRPO against baseline GRPO using two experimental designs reflecting real-world constraints. 
For single-GPU settings (a-d), we compare against vanilla GRPO with the same training batch sizes $(m)$, where $m$ is selected so that one training batch fits within GPU memory. This corresponds to the comparison between the first and the third rows in \cref{fig:method}.
% For single-GPU settings (a-c), we fix the training batch size $(m)$ for both algorithms, simulating the comparison of GRPO and GRPO-PODS in \cref{fig:method}. 
For distributed settings (e-f), we compare against GRPO with gradient accumulation (GRPO-GA), the standard approach for scaling RLVR. In GRPO-GA, large training batches are processed through multiple gradient accumulation steps, enabling updating on larger effective batch sizes at the cost of increased communication overhead and iteration time. We fix the total rollouts generated per prompt $(n)$ and compare GRPO-GA against GRPO-PODS. This corresponds to the comparison between the second and the third rows in \cref{fig:method}.
% For distributed settings (d-e), we fix the number of rollouts generated per prompt $(n)$ and use gradient accumulation to avoid out-of-memory errors, simulating the comparison of GRPO-GA and GRPO-PODS in \cref{fig:method}. 
The detailed hyperparameters used for our experiments are listed in \cref{subsec:exp_hyperparams}.

\cref{fig:main} shows test accuracy over wall-clock training time across all experiment configurations. PODS consistently achieves faster convergence: reaching the baselines' peak accuracies at least $1.7\times$ faster (see \cref{subsec:exp_speedup} for complete results) while often converging to a higher final performance. These results demonstrate PODS' broad applicability across task domains (Math and Chemistry), model scales (3B-7B), architectures (Qwen2.5, Llama3.2), and deployment scenarios, making it a practical improvement for RLVR systems using GRPO.

\subsection{Effect of Rollout and Update Sizes $(n,m)$}
\label{subsec:exp_params}

\begin{figure}[t!]
    \centering
    \begin{subfigure}{\textwidth}
        \centering
        \includegraphics[width=\textwidth]{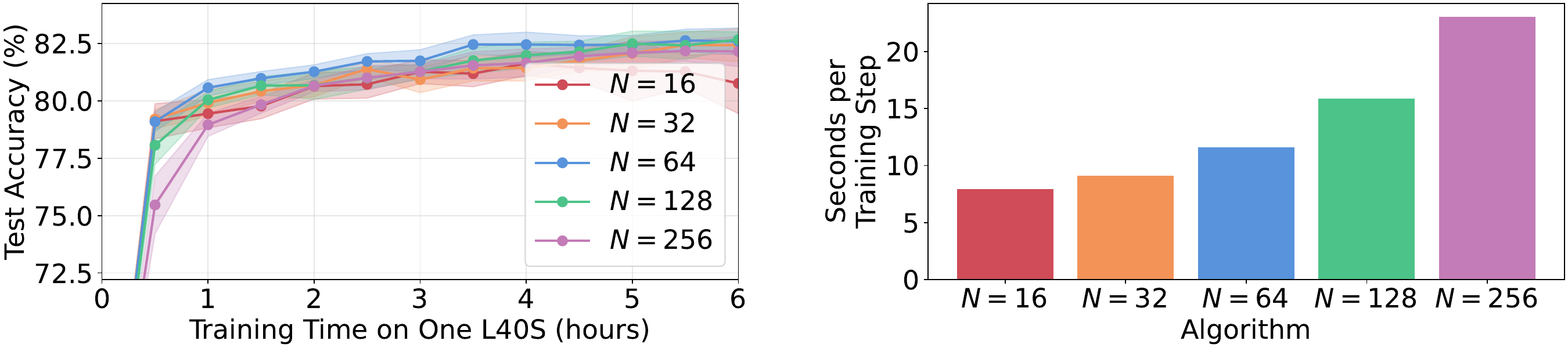}
        \caption{Fixing $m=16$ and varying $n \in \{16, 32, 64, 128, 256\}$}
        \label{subfig:gsm8k_n_scaling}
    \end{subfigure}
    \hfill
    \begin{subfigure}{\textwidth}
        \centering
        \includegraphics[width=\textwidth]{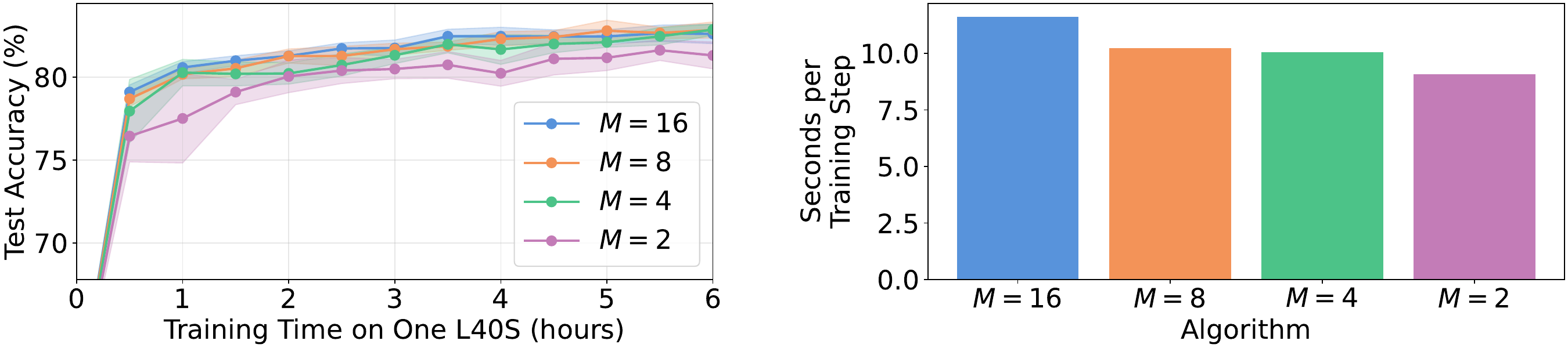}
        \caption{Fixing $n=64$ and varying $m \in \{16, 8, 4, 2\}$}
        \label{subfig:gsm8k_m_scaling}
    \end{subfigure}
    \caption{Performance and per-step run time comparison of GRPO-PODS with max-variance down-sampling across different settings of $n$ and $m$. The training is conducted on the GSM8K dataset with one L40S. For the performance comparison, the x-axis shows the training time, and the y-axis shows the accuracy on the test set. The shaded area represents 1.96 times the standard error of the mean.}
    \label{fig:scaling}
\end{figure}

A key practical question for PODS adoption is how to choose the rollout size $(n)$ and training batch size $(m)$. While a larger $n$ provides more diverse rollouts for selection, it also increases inference costs. Meanwhile, a smaller $m$ reduces update costs but may provide insufficient training signal. As shown in \cref{fig:scaling}, we study these trade-offs on experimental setting (a) to provide deployment guidance.

Increasing rollout size $n$ exhibits diminishing returns with an optimal point around $n=64$. Performance initially improves as larger pools enable better sample selection, but degrades beyond $n=128$ due to two factors: (1) inference runtime grows significantly as GPU memory saturates, and (2) marginal improvements in rollout diversity plateau while computational overhead continues rising.

Training batch size $m$ shows robust performance across a wide range, with minimal degradation until very small values $(m\le 4)$. This suggests PODS' max-variance selection maintains effective learning signals even with aggressive down-sampling ratios up to $16$ where $n=64, m=4$.

\paragraph{Practical guidelines.} These results suggest down-sampling ratio of $2$ to $4$ provides an effective balance of performance and efficiency. For resource-constrained settings, aggressive down-sampling ratios up to $16$ remain viable, while memory-rich environments can benefit from larger rollout pools up to hardware limits.

% We study the effect of the rollout size $n$ and update size $m$ on the performance of GRPO-PODS with max-variance down-sampling in this section. We conduct experiments on the GSM8K dataset with one L40S GPU, and we vary $n$ and $m$ independently. The results are shown in \cref{fig:scaling}. For the rollout size scaling experiment, we fix $m=16$ and vary $n \in \{16, 32, 64, 128, 256\}$. For the update size scaling experiment, we fix $n=64$ and vary $m \in \{16, 8, 4, 2\}$.

% We observe that increasing the rollout size $n$ results in a single-peaked performance curve, where the performance first increases and then decreases as $n$ increases. This can be attributed to two factors: (1) as shown in the right plot of \cref{subfig:gsm8k_n_scaling}, the per-step run time increases as $n$ increases, and such increase is more pronounced when $n$ is large enough to saturate the GPU memory; (2) the overall quality of the rollouts retained for training increases as $n$ increases, but the marginal gain diminishes. Overall, we find that $n=64$ is a good choice for the rollout size when $m=16$.

% For the update size scaling experiment, we observe that decreasing $m$ results in an increased variance of the algorithm's performance, but the overall performance is not significantly affected unless we decrease $m$ to a very small value like $2$ or $4$. This indicates that GRPO-PODS with max-variance down-sampling is robust to the choice of $m$ as long as it is not too small.

\subsection{Comparing Different Down-Sampling Rules}
\label{subsec:exp_rules}

\begin{figure}[h]
    \centering
    \includegraphics[width=\textwidth]{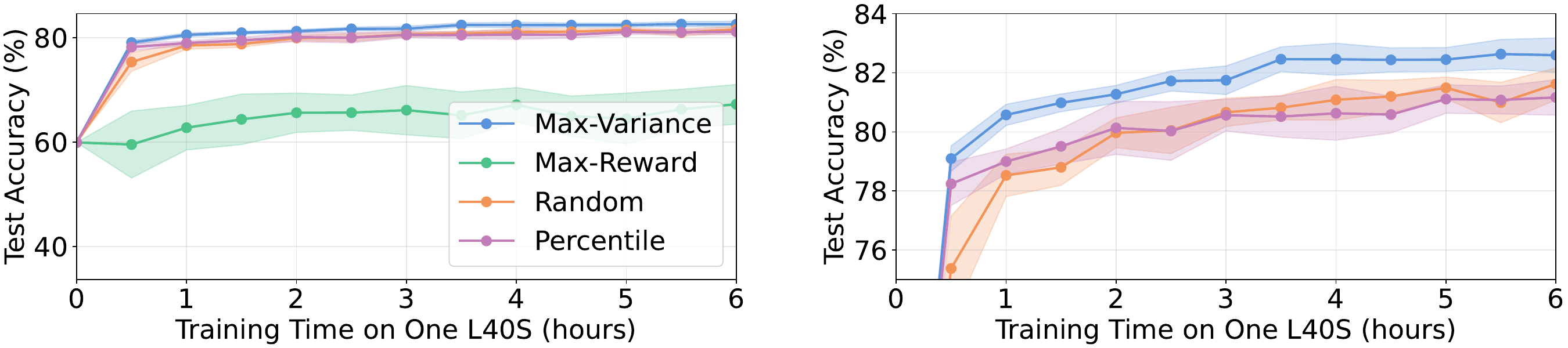}
    \caption{Performance of GRPO-PODS with the max-variance, max-reward, random and percentile down-sampling rules. The training is conducted on the GSM8K dataset with one L40S. The x-axis shows the training time, and the y-axis shows the accuracy on the test set. The shaded area represents 1.96 times the standard error of the mean. The right panel shows a Zoomed-in view of the left one.}
    \label{fig:ds_rules}
\end{figure}

We study the effect of different down-sampling rules on the performance of GRPO-PODS in this section. We conduct experiments on experimental setting (a). We compare four different down-sampling rules: (1) max-variance down-sampling, (2) max-reward down-sampling, (3) random down-sampling, and (4) percentile down-sampling. The results are shown in \cref{fig:ds_rules}. We observe that the max-variance down-sampling rule consistently outperforms the other down-sampling rules throughout the training. This indicates that the max-variance down-sampling rule is effective in selecting informative rollouts for training.

\section{Conclusion and Discussion}
\label{sec:discussions}

We introduced \textbf{PODS}, a lightweight framework that addresses a fundamental bottleneck in modern RLVR training: the asymmetry between embarrassingly parallel rollout generation and memory-intensive policy updates. PODS generates large batches of rollouts in parallel and updates the policy on only an informative subset chosen by the max-variance rule. Our analysis shows that the optimal subset can be found in $O(n \log n)$ time. This simple yet principled approach consistently outperforms standard GRPO under equal wall-clock budgets, delivers at least a $1.7\times$ speedup and reaching higher final accuracy across diverse model architectures, scales, and deployment scenarios. Our ablation study shows that the performance of PODS is robust over a wide range of down-sampling ratios provided $m$ is not too small, empirically confirming our method's efficacy.

% Across multiple dataset, model and hardware configurations, GRPO-PODS consistently outperforms standard GRPO under equal wall-clock budgets, converging faster and reaching higher final accuracy. Our ablation study shows that the performance of PODS is robust over a wide range of down-sampling ratios provided $m$ is not too small, empirically confirming our theoretical motivation.

% \paragraph{Limitations.} Our evaluation focuses on mathematical-reasoning tasks with verifiable rewards. Other domains such as open-ended dialogue or code generation may exhibit distinct dynamics of the algorithms. Moreover, in workloads that demand greater prompt diversity, similar gains might be obtained by processing more prompts per iteration with fewer rollouts per prompt and accumulating gradients across prompts---an alternative path to address the inference-update asymmetry. Finally, because PODS alters the training rollout distribution through selective down-sampling, it behaves off-policy and may be unsuitable when strict on-policy guarantees are required.

\paragraph{Limitations.} As we emphasize in the paper, our evaluation focuses on RLVR tasks where response correctness is verifiable. Other domains such as open-ended dialogue may exhibit distinct dynamics of the algorithms. Moreover, in workloads that demand greater prompt diversity, similar gains might be obtained by processing more prompts per iteration with fewer rollouts per prompt and accumulating gradients across prompts, which is an alternative path to address the inference-update asymmetry. Finally, because PODS alters the training
rollout distribution through selective down-sampling, it behaves off-policy and may be unsuitable when strict
on-policy guarantees are required despite its improved empirical performance.

\textbf{Scope of claims.} Our empirical and theoretical claims in this paper are specific to PODS instantiated within GRPO. While we expect the inference-update asymmetry to be present in other RLVR pipelines, demonstrating that down-sampling yields comparable benefits for other RL methods (e.g., PPO-style objectives or value-based variants) requires separate analysis and experimentation.

\paragraph{Discussion and future work.} PODS is a general framework that admits different down-sampling rules. In this paper, we focus on the max-variance rule due to its superior empirical performance in common RLVR settings. In some scenarios, other rules might be more effective. For example, when different prompts cause the model to have highly varying reward distributions, applying the max-variance rule across all rollouts may lead to over-sampling from a small subset of prompts with extreme difficulty levels. In such cases, applying the max-variance rule within each prompt and then selecting a balanced subset across prompts may be more effective. More broadly, the down-sampling rule can be designed to take into account more information beyond the reward values, such as the rollouts' entropy, similarity measures between rollouts, or even a target reward distribution that we wish to down-sample towards. Exploring these alternative down-sampling rules and their empirical performance across different settings is an interesting direction for future work.

% \paragraph{Ethics statement.} We anticipate our work will primarily have positive social impact by improving the computational efficiency and effectiveness of RL training for LLMs, potentially democratizing access to high-quality reasoning models. However, by lowering the computational barriers to training powerful reasoning systems, our method may accelerate capabilities that could be misused. This heightens the importance of responsible release practices to mitigate harmful behaviors. Our open-source release of code and experimental frameworks aims to facilitate reproducibility while encouraging informed and safe adoption within the research community.

% \paragraph{Reproducibility statement.} We describe the rewards used in \cref{subsec:rewards}, list the key hyperparameters in \cref{subsec:exp_hyperparams}, and an anonymized version of the code used to run the experiments in this paper is attached to our submission as supplementary material on OpenReview. We will publicly release the code on GitHub, and include its link in the next version of our paper. We note that all of the datasets used in this paper are open-source, and the models we use are all open-weight and available publicly on HuggingFace.

% \paragraph{LLM usage statement.} In this work, we used LLMs as an assist tool in polishing the language of our writing in the paper and auto-completing some of our evaluation code.

% \begin{ack}
%     This work is supported in part by NSF IIS-2200410.
% \end{ack}

\bibliographystyle{tmlr}
\bibliography{ref}

\appendix

\newpage

\section{Additional Experimental Details}
\label{app:exp_details}

\subsection{Reward Functions}
\label{subsec:rewards}
We list the reward functions we use in our experiments below.

\textbf{Accuracy ($\mathbf{1}$ for correct, $\mathbf{0}$ for incorrect):}
For mathematical reasoning tasks, we use a binary reward indicating whether the final answer is correct. This is computed by comparing the model's final answer to the ground truth solution using \LaTeX\ parsing and symbolic verification, which allows for robust equivalence checking even when the model's answer is not in the exact same format as the ground truth. For Chemistry problems, the answer is always a letter in \{A, B, C, D\}, so we can directly compare with the correct answer.

\textbf{Format ($\mathbf{1}$ for compliant, $\mathbf{0}$ for non-compliant):} For a response to be considered format-compliant, it must follow the exact XML structure we specify for reasoning and answer. This requires reasoning to be enclosed within \texttt{<think>} tags and the final answer within \texttt{<answer>} tags, following the exact pattern \verb|<think>\n...\n</think>\n<answer>\n...\n</answer>|.

\textbf{Tag count ($\mathbf{0}$ to $\mathbf{1}$ partial credit):} The model receives 0.25 points each for correct placement of \verb|<think>\n|, \verb|\n</think>\n|, \verb|\n<answer>\n|, and \verb|\n</answer>| tags. We allow partial credit for partially correct formatting.

% These rewards encourage both mathematical accuracy and consistent structured reasoning, with the tag count reward providing more nuanced feedback than the binary format reward during training.

\subsection{Hyperparameters}
\label{subsec:exp_hyperparams}

In \cref{tab:hyperparameters}, we list the key hyperparameters we use for different experimental settings.

\begin{table}[h]
    \centering
    \caption{Hyperparameters for different experimental settings.}
    \label{tab:hyperparameters}
    \begin{tabular}{lcccccc}
        \toprule
        \textbf{Setting} & \textbf{(a)} & \textbf{(b)} & \textbf{(c)} & \textbf{(d)} & \textbf{(e)} & \textbf{(f)} \\
        \midrule
        Optimizer & AdamW & AdamW & AdamW & AdamW & AdamW & AdamW \\
        Max Sequence Length & $1024$ & $1024$ & $1024$ & $1024$ & $2048$ & $2048$ \\
        Lora Rank & $64$ & $64$ & $64$ & $64$ & N/A & N/A \\
        Lora Alpha & $64$ & $64$ & $64$ & $64$ & N/A & N/A \\
        KL Coefficient & $0.00$ & $0.04$ & $0.00$ & $0.00$ & $0.00$ & $0.00$ \\
        Learning Rate & $5\cdot 10^{-6}$ & $2\cdot 10^{-6}$ & $5\cdot 10^{-6}$ & $5\cdot 10^{-6}$ & $2\cdot 10^{-5}$ & $1.5\cdot 10^{-5}$ \\
        Weight Decay & $0.1$ & $0.1$ & $0.1$ & $0.1$ & $0.1$ & $0.1$ \\
        Grad Clipping & $1.0$ & $1.0$ & $1.0$& $1.0$ & $1.0$ & $1.0$ \\
        \midrule
        GA Steps (GRPO-PODS) & $1$ & $1$ & $1$ & $1$ & $4$ & $4$ \\
        Rollout Batch Size (GRPO-PODS) & $64$ & $64$ & $32$ & $64$ & $128$ & $128$ \\
        Update Batch Size (GRPO-PODS) & $16$ & $16$ & $8$ & $16$ & $32$ & $32$ \\
        Effective $n$ (GRPO-PODS) & $64$ & $64$ & $32$ & $64$ & $512$ & $512$ \\
        Effective $m$ (GRPO-PODS) & $16$ & $16$ & $8$ & $16$ & $128$ & $128$ \\
        Down-Sampling Ratio & $4$ & $4$ & $4$ & $4$ & $4$ & $4$ \\
        \midrule
        GA Steps (GRPO) & $1$ & $1$ & $1$ & $1$ & N/A & N/A \\
        Rollout Batch Size (GRPO) & $16$ & $8$ & $16$ & $16$ & N/A & N/A \\
        Update Batch Size (GRPO) & $16$ & $8$ & $16$ & $16$ & N/A & N/A \\
        Effective $n$ (GRPO) & $16$ & $8$ & $16$ & $16$ & N/A & N/A \\
        Effective $m$ (GRPO) & $16$ & $8$ & $16$ & $16$ & N/A & N/A \\
        \midrule
        GA Steps (GRPO-GA)           & N/A & N/A & N/A & N/A & $16$ & $16$ \\
        Rollout Batch Size (GRPO-GA) & N/A & N/A & N/A & N/A & $32$ & $32$ \\
        Update Batch Size (GRPO-GA)  & N/A & N/A & N/A & N/A & $32$ & $32$ \\
        Effective $n$ (GRPO-GA)      & N/A & N/A & N/A & N/A & $512$ & $512$ \\
        Effective $m$ (GRPO-GA)      & N/A & N/A & N/A & N/A & $512$ & $512$ \\
        \bottomrule
    \end{tabular}
\end{table}

\paragraph{Note on gradient accumulation.} For experiment settings (e) and (f), we ensure a fair comparison between GRPO-PODS and GRPO-GA by matching the total number of rollouts (effective $n$) generated per prompt. This is done by equating the product of rollout batch size and GA steps across both methods. In the \texttt{open-r1} implementation, GA steps determine both rollout generation and training updates. For example, with rollout batch size $128$ and GA steps $4$, the effective $n$ is $128 \times 4 = 512$. We fixed this number at $512$ for both GRPO-PODS and GRPO-GA. For GRPO-PODS, each rollout batch is down-sampled by a factor of $4$, resulting in an update batch size of $32$ and an effective $m$ of $32 \times 4 = 128$. Because down-sampling is applied directly after generating each batch rather than after aggregation, GRPO-GA must increase GA steps by $4\times$ (from $4$ to $16$) to maintain the same effective $n$. This adjustment ensures that both variants process an equal number of rollouts while respecting their structural differences.
\subsection{On Advantage Normalization}
\label{subsec:exp_adv_norm}

\begin{figure}[h]
    \centering
    \includegraphics[width=\textwidth]{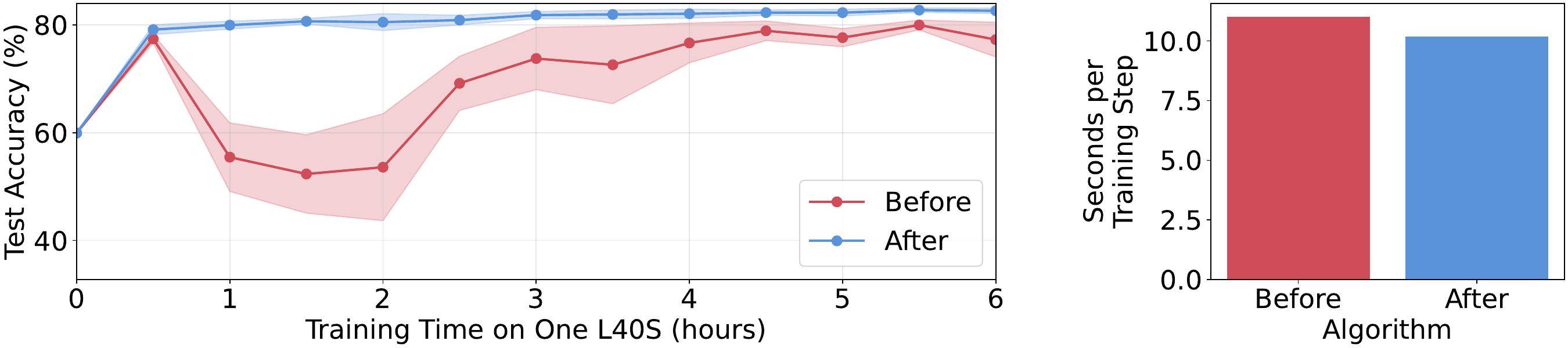}
    \caption{Performance and per-step run time comparison of the two design choices of advantage normalization in GRPO-PODS with max-variance down-sampling in experimental setting (a). ``After'' refers to computing the advantage normalization statistics using the down-sampled batch, while ``Before'' refers to doing so using the full rollout batch before down-sampling. The x-axis shows the training time, and the y-axis shows the accuracy on the test set. The shaded area represents 1.96 times the standard error of the mean.}
    \label{fig:adv_norm} 
\end{figure}

In \cref{subsec:multigrpo}, we mention that in GRPO-PODS, we first apply the down-sampling to the rollout batch, and then compute the advantage normalization statistics (mean and standard deviation) using the down-sampled batch. This design ensures the advantage normalization is consistent with the actual data used for policy updates, so that each update batch has a total advantage of $0$. In this section, we study the effect of this design choice by comparing it with an alternative design where the advantage normalization statistics are computed using the full rollout batch before down-sampling. We conduct this comparison in the experimental setting (a) where we train Qwen2.5 (3B) on GSM8K with one L40S GPU. The results are shown in \cref{fig:adv_norm}. We observe that computing the advantage normalization statistics using the down-sampled batch leads to better performance than using the full rollout batch. We hypothesize that the fact that the advantage normalization is consistent with the actual data used for policy updates is essential to the improved performance.

\subsection{PODS' Speed Up Ratio Over GRPO}
\label{subsec:exp_speedup}

In \cref{fig:main}, we observe that GRPO-PODS consistently outperforms GRPO in terms of performance as the training proceeds. For each of the six plots in \cref{fig:main}, we compute the speed up ratio of GRPO-PODS over GRPO, i.e., the ratio between the time taken by GRPO and that taken by GRPO-PODS to reach $0.99\times$ the peak performance of GRPO. The results are shown in \cref{tab:speedup}. We observe that our method achieves a speed up ratio between $1.7\times$ and $3.0\times$ over GRPO across the settings.

\begin{table}[h]
    \centering
    \caption{Speed up ratio of GRPO-PODS over GRPO in \cref{fig:main}.}
    \label{tab:speedup}
    \begin{tabular}{lcccccc}
        \toprule
        \textbf{Setting} & \textbf{(a)} & \textbf{(b)} & \textbf{(c)} & \textbf{(d)} & \textbf{(e)} & \textbf{(f)} \\
        \midrule
        Speed Up Ratio & $2.0\times$ & $3.0\times$ & $2.0\times$ & $1.8\times$ & $1.7\times$ & $1.7\times$ \\
        \bottomrule
    \end{tabular}
\end{table}

\newpage
\subsection{Additional Evaluation for Experimental Settings (a) and (b)}
\label{subsec:exp_additional_eval}

\begin{figure}[h]
    \centering
    \begin{subfigure}{\textwidth}
        \centering
        \includegraphics[width=\textwidth]{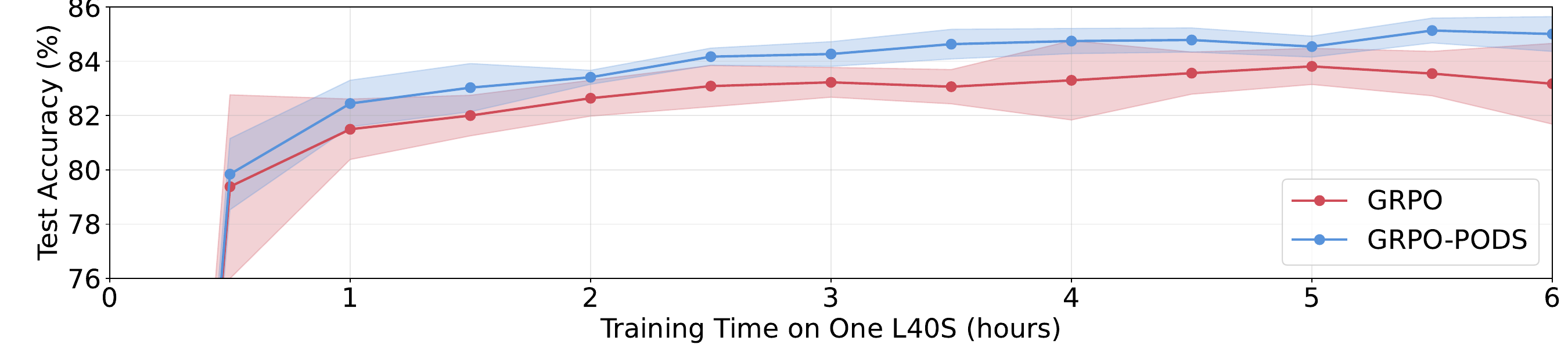}
        \caption{Training Qwen2.5 (3B) on GSM8K with one L40S GPU and Evaluating on GSM8K Platinum}
        \label{subfig:gsm8k_gsm8k_platinum}
    \end{subfigure}
    \hfill
    \begin{subfigure}{\textwidth}
        \centering
        \includegraphics[width=\textwidth]{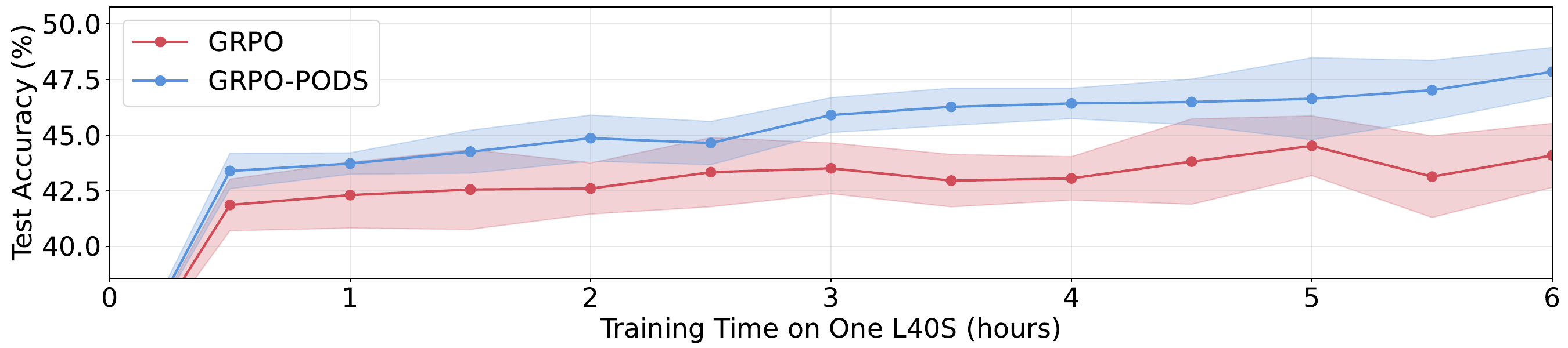}
        \caption{Training Qwen2.5 (3B) GSM8K with one L40S GPU and Evaluating on MATH}
        \label{subfig:gsm8k_math}
    \end{subfigure}
    \hfill
    \begin{subfigure}{\textwidth}
        \centering
        \includegraphics[width=\textwidth]{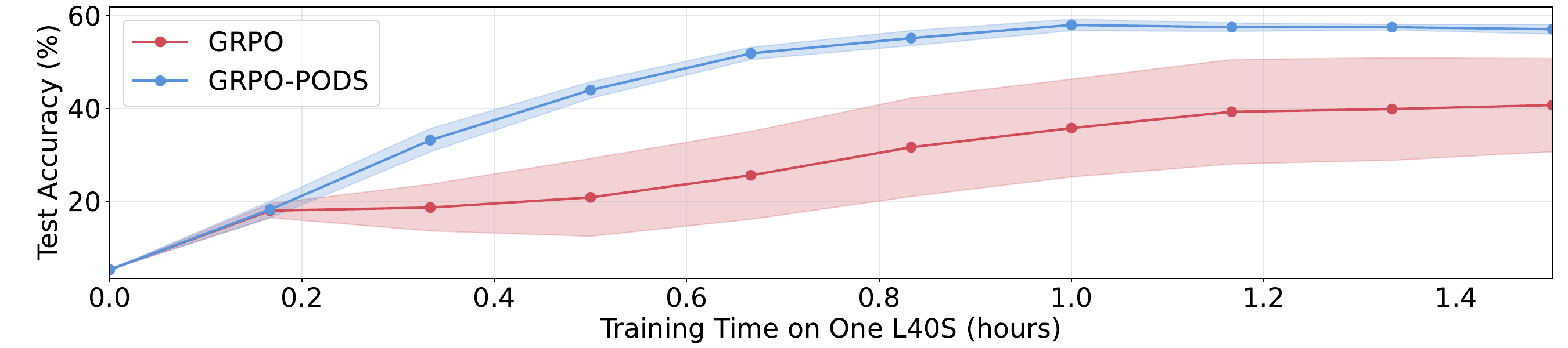}
        \caption{Training Llama3.2 (3B) on GSM8K with one L40S GPU and Evaluating on GSM8K Platinum}
        \label{subfig:llama_gsm8k_platinum}
    \end{subfigure}
    \hfill
    \begin{subfigure}{\textwidth}
        \centering
        \includegraphics[width=\textwidth]{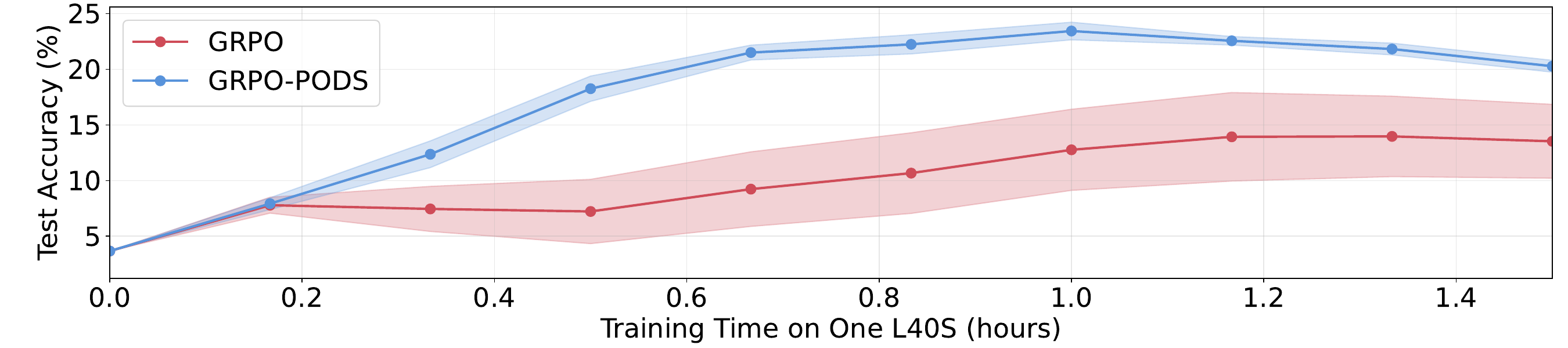}
        \caption{Training Llama3.2 (3B) on GSM8K with one L40S GPU and Evaluating on MATH}
        \label{subfig:llama_math}
    \end{subfigure}
    \caption{Additional performance comparison of standard GRPO and GRPO-PODS with max-variance down-sampling in experimental settings (a-b). The x-axis shows the training time, and the y-axis shows the accuracy on the test set. The shaded area represents 1.96 times the standard error of the mean.}
    \label{fig:additional_eval} 
\end{figure}

To demonstrate that the performance improvement of GRPO-PODS over GRPO is not limited to the specific test set used in \cref{fig:main}, we evaluate the trained models on additional test sets for experimental settings (a) and (b). For both settings, we evaluate the trained models on the GSM8K Platinum test set, which is a contamination-resistant subset of GSM8K, and MATH. The results are shown in \cref{fig:additional_eval}. We observe that GRPO-PODS consistently outperforms GRPO across all the test sets, demonstrating the robustness of the performance improvement, and indicating the improvement is not due to overfitting to a specific test set.

\newpage
\subsection{Average Completion Length Over Time}
\label{subsec:exp_length}

\begin{figure}[H]
    \centering
    \begin{subfigure}{\textwidth}
        \centering
        \includegraphics[width=\textwidth]{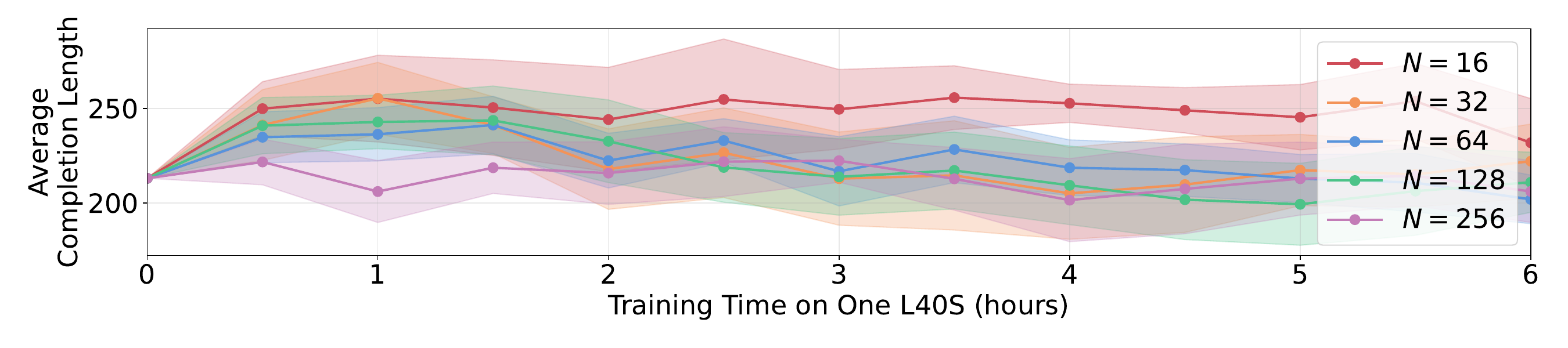}
        \caption{Fixing $m=16$ and varying $n \in \{16, 32, 64, 128, 256\}$}
        \label{subfig:gsm8k_n_scaling_completion}
    \end{subfigure}
    \hfill
    \begin{subfigure}{\textwidth}
        \centering
        \includegraphics[width=\textwidth]{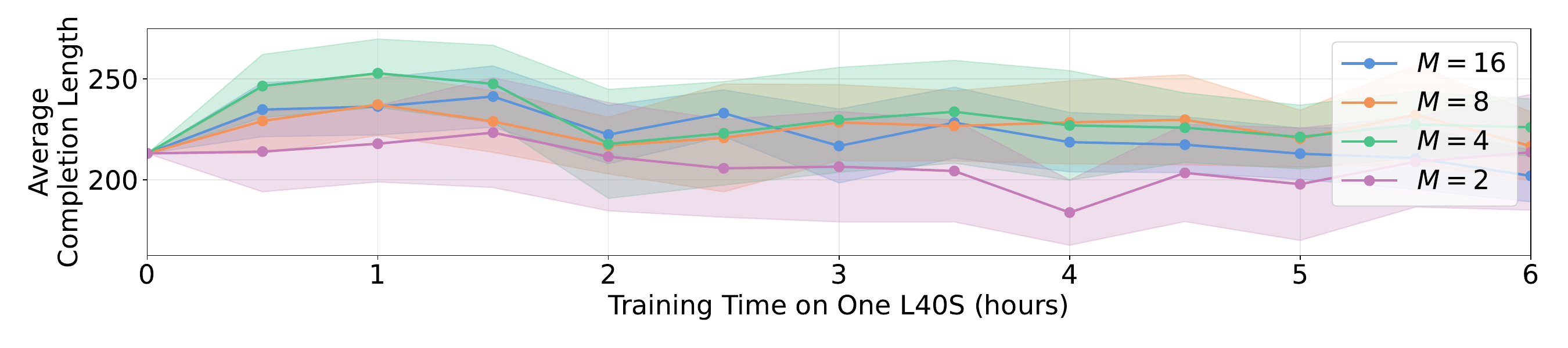}
        \caption{Fixing $n=64$ and varying $m \in \{16, 8, 4, 2\}$}
        \label{subfig:gsm8k_m_scaling_completion}
    \end{subfigure}
    \caption{Average completion length over time of the trained models in \cref{subsec:exp_params}'s experiments. The x-axis shows the training time, and the y-axis shows the average completion length in tokens. The shaded area represents 1.96 times the standard error of the mean.}
    \label{fig:scaling_completion}
\end{figure}

\begin{figure}[H]
    \centering
    \includegraphics[width=\textwidth]{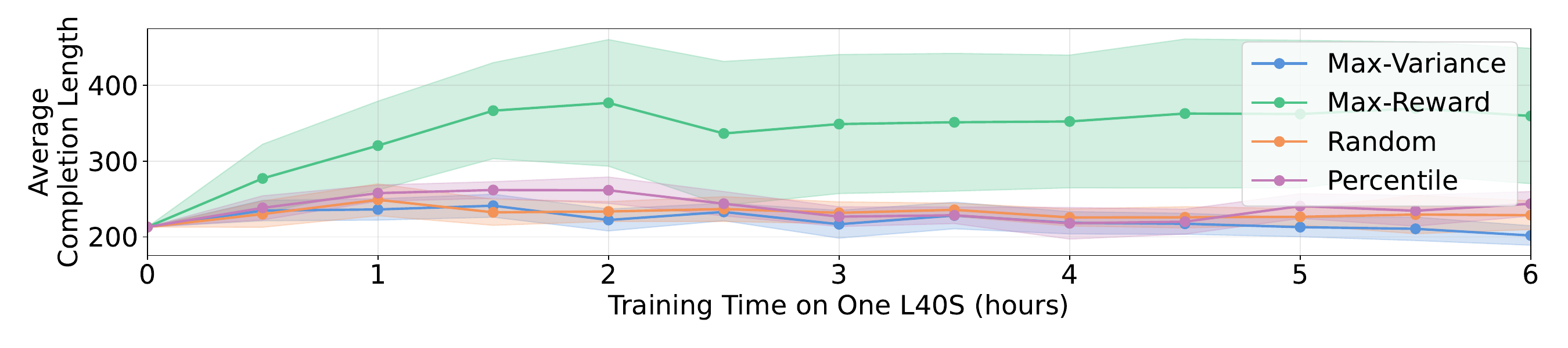}
    \caption{Average completion length over time of the trained models in \cref{subsec:exp_rules}'s experiments. The x-axis shows the training time, and the y-axis shows the average completion length in tokens. The shaded area represents 1.96 times the standard error of the mean.}
    \label{fig:ds_rules_completion}
\end{figure}

\begin{figure}[ht]
    \centering
    \begin{subfigure}{\textwidth}
        \centering
        \includegraphics[width=0.9\textwidth]{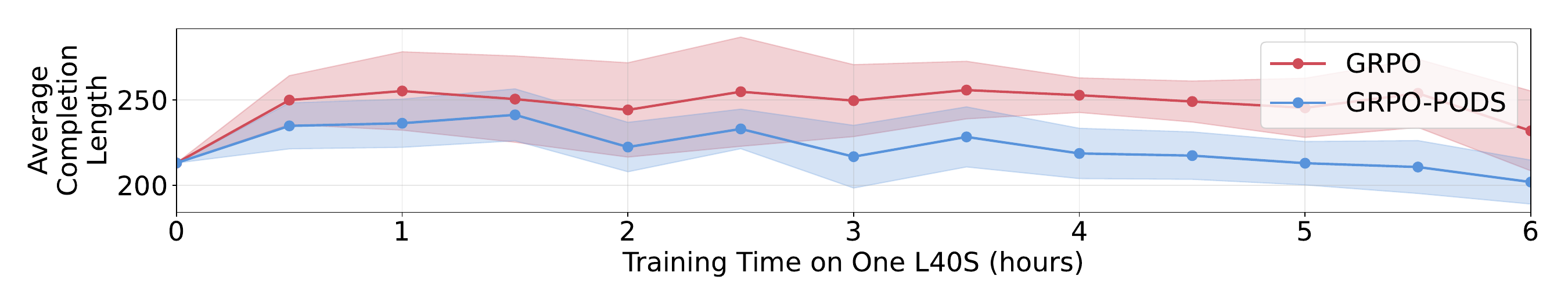}
        \caption{Training Qwen2.5 (3B) on GSM8K with one L40S GPU}
        \label{subfig:gsm8k_main_completion}
    \end{subfigure}
    \hfill
    \begin{subfigure}{\textwidth}
        \centering
        \includegraphics[width=0.9\textwidth]{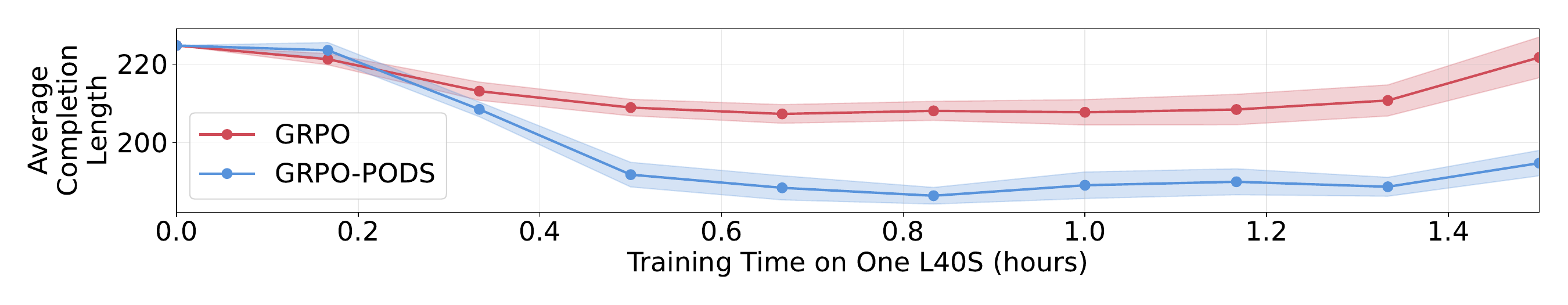}
        \caption{Training Llama3.2 (3B) on GSM8K with one L40S GPU}
        \label{subfig:llama_main_completion}
    \end{subfigure}
    \hfill
    \begin{subfigure}{\textwidth}
        \centering
        \includegraphics[width=0.9\textwidth]{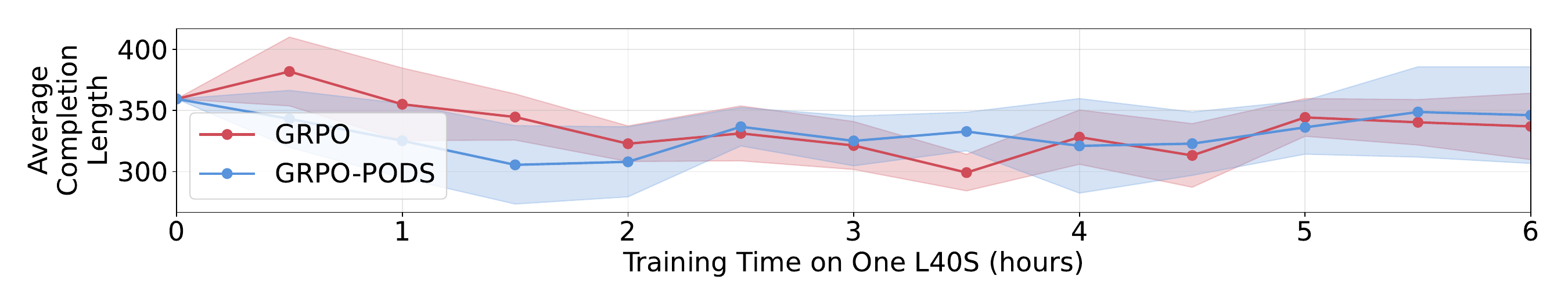}
        \caption{Training Qwen2.5 (3B) on MATH with one L40S GPU}
        \label{subfig:math_main_completion}
    \end{subfigure}
    \hfill
    \begin{subfigure}{\textwidth}
        \centering
        \includegraphics[width=0.9\textwidth]{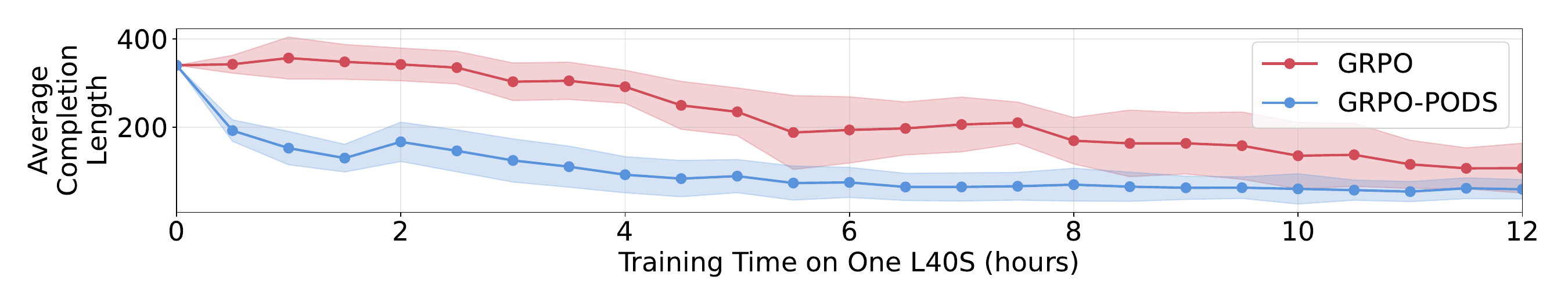}
        \caption{Training Qwen2.5 (3B) on Chemistry with one L40S GPU}
        \label{subfig:schiknow_main_completion}
    \end{subfigure}
    \hfill
    \begin{subfigure}{\textwidth}
        \centering
        \includegraphics[width=0.9\textwidth]{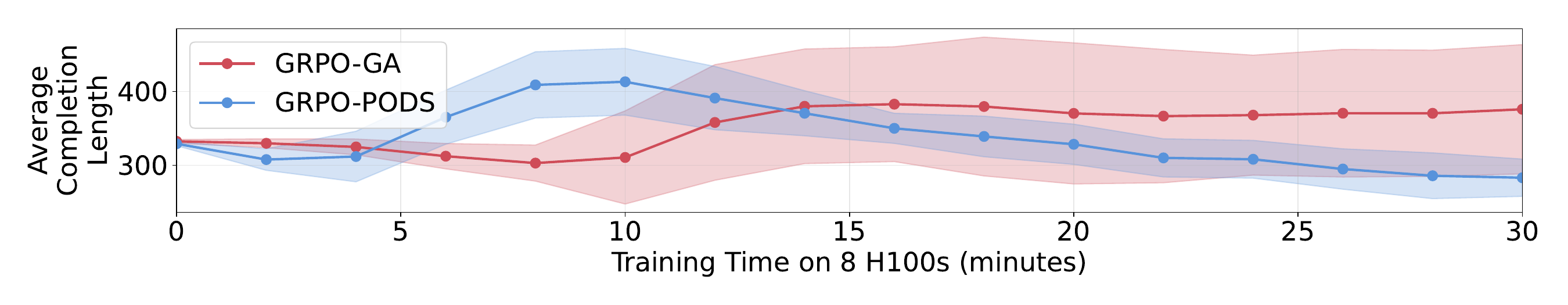}
        \caption{Training Qwen2.5 (3B) on GSM8K with 8 H100 GPUs}
        \label{subfig:gsm8k_scale_completion}
    \end{subfigure}
    \hfill
    \begin{subfigure}{\textwidth}
        \centering
        \includegraphics[width=0.9\textwidth]{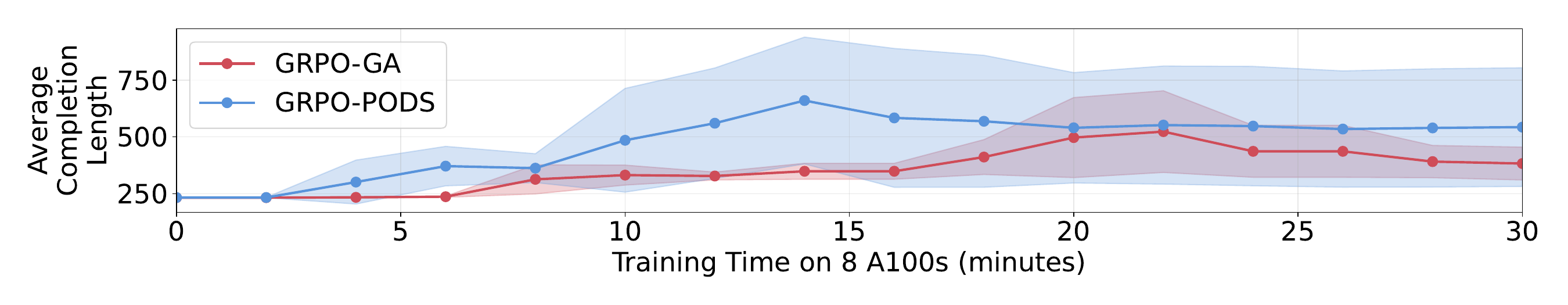}
        \caption{Training Qwen2.5 (7B) on GSM8K with 8 A100 GPUs}
        \label{subfig:gsm8k_scale_7b_completion}
    \end{subfigure}
    \\
    \caption{Average completion length over time of the trained models in \cref{subsec:exp_grpo}'s experiments. The x-axis shows the training time, and the y-axis shows the average completion length in tokens. The shaded area represents 1.96 times the standard error of the mean.}
    \label{fig:main_completion}
\end{figure}

We include additional evaluation of the average completion length over the training time for each of the experiments we conduct in \cref{sec:experiments}. We present the average completion length  results in \cref{fig:main_completion,fig:scaling_completion,fig:ds_rules_completion}, in correspondence to \cref{fig:main,fig:scaling,fig:ds_rules} respectively. In most of the experimental settings, we observe that the average completion length stays relatively stable over the training time.

\end{document}